\documentclass{article}
\pdfoutput=1

\PassOptionsToPackage{numbers, compress}{natbib}
\PassOptionsToPackage{dvipsnames}{xcolor}
\usepackage[final]{neurips_2021}

\usepackage[utf8]{inputenc} 
\usepackage[T1]{fontenc}    
\usepackage{hyperref}       
\usepackage{url}            
\usepackage{booktabs}       
\usepackage{amsfonts}       
\usepackage{nicefrac}       
\usepackage{microtype}      
\usepackage{xcolor}         
\usepackage{multirow}
\usepackage{multicol}
\usepackage{makecell}
\usepackage{amsmath}
\usepackage{booktabs}
\usepackage[shortlabels]{enumitem}
\usepackage{graphicx}
\usepackage{amsthm}
\theoremstyle{definition}
\newtheorem{definition}{Definition}

\title{On the Practical Consistency of \\Meta-Reinforcement Learning Algorithms}

\author{%
    Zheng Xiong, Luisa Zintgraf, Jacob Beck, Risto Vuorio, Shimon Whiteson \\
    Department of Computer Science \\
    University of Oxford \\
    \texttt{zheng.xiong@cs.ox.ac.uk} \\
}

\begin{document}

\maketitle

\begin{abstract}
Consistency is the theoretical property of a meta learning algorithm that ensures that, under certain assumptions, it can adapt to any task at test time. 
An open question is whether and how theoretical consistency translates into practice, in comparison to inconsistent algorithms.
In this paper, we empirically investigate this question on a set of representative meta-RL algorithms.
We find that theoretically consistent algorithms can indeed usually adapt to out-of-distribution (OOD) tasks, while inconsistent ones cannot, although they can still fail in practice for reasons like poor exploration. 
We further find that theoretically inconsistent algorithms can be made consistent by continuing to update all agent components on the OOD tasks, and adapt as well or better than originally consistent ones. 
We conclude that theoretical consistency is indeed a desirable property, and inconsistent meta-RL algorithms can easily be made consistent to enjoy the same benefits.
\end{abstract}

\section{Introduction}

For intelligent agents, picking up new tasks is of great practical importance: the real world is noisy and constantly changing, and an agent should reuse its prior knowledge effectively in order to learn new skills.
However, the ability of reinforcement learning (RL) agents to do so is still quite limited due to sample inefficiency.
Meta-reinforcement learning (meta-RL) algorithms aim to alleviate this problem by meta-training on a set of related tasks to learn a good \emph{inductive bias} on the types of tasks the agent might encounter, which enables fast adaptation to similar (meta-test) tasks. 

While a strong inductive bias helps the agent adapt quickly to tasks similar to those seen during meta-training, the agent may be less  adaptive to out-of-distribution (OOD) tasks \citep{botvinick2019reinforcement}. 
In fact, it may take an agent longer to adapt than if it had a weaker inductive bias. 
In a constantly changing world, an agent will likely need to solve OOD tasks that it has not seen before.
Consequently, it is desirable for meta-RL algorithms to be able to adapt to OOD tasks beyond their meta-training distribution. 

A related concept is \emph{consistency}: a meta learning algorithm is consistent if it can -- in theory, under certain assumptions, in the limit -- find the optimal solution for any test task \citep{finn2018learning}. 
A typical class of consistent meta-RL algorithms are gradient-based methods like MAML \citep{finn2017model}, E-MAML \citep{stadie2018some}, and ProMP \citep{rothfuss2018promp}. 
A typical class of \emph{inconsistent} meta-RL algorithms are context-based methods, such as RL$^2$ \citep{duan2016rl,wang2016learning}, PEARL \citep{rakelly2019efficient}, and VariBAD \citep{zintgraf2019varibad}.
The reason gradient-based methods are consistent is because that they adapt using gradient descent at test time, and gradient descent is known to be consistent \citep{zhang2017understanding}. 
On the other hand, the inconsistent context-based methods typically consist of a trajectory encoder and a policy which are meta-trained on a given task distribution via gradient descent, but fixed at meta-test time without further gradient adaptation. 
Therefore we have no guarantees that the adaptation procedure they learn on a specific task distribution is going to be consistent on arbitrary test task.

Due to its close relationship to OOD adaptation, consistency is considered a desirable property \citep{finn2018learning,mendonca2020meta,mitchell2021offline}. 
However, it remains a \emph{theoretical} property that only holds true under certain assumptions such as infinite adaptation data, or consistency of policy gradients. 
As these assumptions usually do not hold empirically, it remains an open question how theoretical consistency translates into ``practical consistency'', i.e., the ability of a meta-RL algorithm to adapt to OOD tasks given a large but limited adaptation budget. 
In this paper, we aim to shed some light on this question in the context of meta-RL.

We experiment with three representative meta-RL algorithms, MAML \citep{finn2017model}, RL$^2$ \citep{duan2016rl,wang2016learning}, and VariBAD \citep{zintgraf2019varibad}, on a set of diverse tasks, ranging from simple 2D navigation to complex continuous control in Mujoco \citep{todorov2012mujoco}, with different amounts of distribution drift. 
Our main findings are:
\begin{enumerate}[topsep=0pt]
\itemsep0em 
    \item In most cases, theoretical consistency provides a good indication of practical consistency, i.e., consistent methods can reasonably adapt to OOD tasks, while inconsistent ones cannot. 
    \item Theoretically inconsistent algorithms can be made consistent and adapt to OOD tasks if we continue training all agent components on the meta-test tasks via gradient adaptation. 
    Empirically, this strategy achieves similar or even better adaptation performance compared to the originally consistent methods. 
    \item Theoretical consistency is not always sufficient for practical consistency. We identify one such failure case where the agent fails to solve OOD tasks due to poor exploration. 
\end{enumerate}
In conclusion, we believe that theoretical consistency is indeed a desirable property, and inconsistent meta-RL algorithms can easily be made consistent. 
We also argue that a strong baseline for any study on OOD adaptation is a current state-of-the-art in-distribution meta-RL algorithm which is adapted OOD by continuing to update the meta-learned policy on the meta-test task(s). 
\section{Background and Related Work}

\subsection{Formulation of Meta-RL}
\label{sec:bg:meta_rl}

We define a Markov Decision Process (MDP) as a tuple $ M = \left( \mathcal{S}, \mathcal{A}, R, T, \gamma, H \right)$, where $\mathcal{S}$ is the state space, $\mathcal{A}$ the action space, $R(r_{t+1}|s_t,a_t,s_{t+1})$ the reward function, $T(s_{t+1}|s_t,a_t)$ the transition function, $\gamma$ the discount factor, and $H$ the horizon (we omit the initial state transition here for simplicity of notation). The goal of standard RL is to learn a policy $\pi$ that maximizes the expected return $J(\pi) = \mathbb{E}_{T,\pi} \left[ \sum_{t=0}^{H-1} \gamma^t R(r_{t+1}|s_t,a_t,s_{t+1}) \right]$. 

For meta-RL, we are given a task distribution $p(M)$ over MDPs, where each $M_i \sim p(M)$ is defined as $ M_i = \left( \mathcal{S}, \mathcal{A}, R_i, T_i, \gamma, H \right)$. 
The objective is to maximise learning performance in a randomly sampled task with initially unknown reward and transition function.\footnote{The exact objective may differ between meta-RL methods, so we do not define this formally here. In our experiments, we follow the respective algorithm's default choices.}
We distinguish between the meta-train distribution $p_{\text{train}}$, and the meta-test distribution $p_{\text{test}}$. 
In this paper we are interested in adaptation to out-of-distribution (OOD) tasks, i.e., 
tasks $M$ for which $p_{\text{train}}(M) = 0$ and $p_{\text{test}}(M) > 0$. 

\subsection{Theoretical Consistency of Meta Learning Algorithms}
\label{subsec:theoretical-consistency}

\citet{finn2018learning} defines a learning algorithm as a function that processes data $\mathcal{D}$ collected from a task $\mathcal{T}$ to make predictions $y^\star$ from new inputs $x^\star$, and considers a learning algorithm to be consistent if it can find the true function mapping from $x^\star$ to $y^\star$ when provided infinite data, i.e., 
\begin{equation}
    \lim_{|\mathcal{D}| \rightarrow \infty} f(\mathcal{D}, x^\star) \rightarrow y^\star \quad \forall x^\star, y^\star \in \mathcal{T}. 
\end{equation}
In meta-RL, $f$ is often a policy (of which there may be multiple optimal ones) or a value function.

Meta-learning algorithms that adapt via stochastic gradient descent (SGD) are often called consistent \citep{finn2018learning,mendonca2020meta,mitchell2021offline}, citing the empirical evidence that SGD is consistent in practice \citep{zhang2017understanding}.

We adopt this convention in this paper when categorising algorithms as consistent, but note that deep reinforcement learning algorithms are typically not theoretically consistent.
This again motivates our question of whether meta-RL algorithms that are labelled as consistent truly have practical advantages compared to inconsistent methods.

\subsection{Meta-RL Algorithms}
\label{subsec:meta-rl-algos}

Existing meta-RL algorithms can be categorized according to what is meta-learned. 
We focus on methods that learn adaptive \emph{policies}, i.e.,  policy networks capable of few-shot adaptation to new tasks, as opposed to meta-learning things like loss functions \citep{houthooft2018evolved,kirsch2019improving,oh2020discovering}, hyperparameters \citep{xu2018meta,zahavy2020self}, or environments \citep{gupta2018unsupervised,dennis2020emergent}.
The meta-RL algorithms investigated in this paper can be further categorized as gradient-based or context-based. 
Gradient-based methods \citep{finn2017model,stadie2018some,rothfuss2018promp,zintgraf2019fast,mendonca2020meta} adapt to new tasks via gradient updates.
These algorithms often differ by network architecture, which parameters are updated, or which loss function is used.
Context-based methods \citep{duan2016benchmarking,wang2016learning,rakelly2019efficient,zintgraf2019varibad,fakoor2019meta} aim to infer a latent context per task, and learn a universal policy that conditions on the inferred context. 
These algorithms often differ by how they encode the context, and how the agent uses this context to act.
We experiment with three representative meta-RL algorithms and test their practical consistency:

\textbf{MAML} \citep{finn2017model} is a gradient-based method that meta-learns a good model initialization to enable fast adaptation. It adapts to meta-test tasks via policy gradients starting from the meta-learned initialization. MAML is considered to be a consistent method.

\textbf{RL$^2$} \citep{duan2016rl,wang2016learning} is a context-based method, where the policy is a recurrent neural network (RNN) that conditions on the state, observed rewards, and taken actions.
At test time, no gradients are taken: adaptation happens via the RNN, which has learned to infer the task via its hidden state.
RL$^2$ is considered to be inconsistent, as there is no guarantee that the RNN can encode OOD tasks, or that the RNN's fixed parameters could interpret the OOD hidden state correctly, even with infinite data.

\textbf{VariBAD} \citep{zintgraf2019varibad} is a context-based method. Unlike RL$^2$, VariBAD decouples learning how to do task inference from policy training. In VariBAD, a VAE \citep{kingma2014auto} is trained to predict environment dynamics given the agent's experience. The VAE's latent state represents the task belief and is passed to the policy.
VariBAD is considered to be inconsistent for the same reasons as RL$^2$.

\subsection{Existing Approaches for OOD Adaptation}

Although most meta-RL algorithms are designed for in-distribution adaptation, there are also several that explicitly consider OOD adaptation by reusing meta-training data, e.g., via experience relabelling \citep{mendonca2020meta} or sample reweighting \citep{fakoor2019meta}. 
MIER \citep{mendonca2020meta} only updates the task inference module via gradient descent for in-distribution adaptation, and additionally updates the policy for out-of-distribution adaptation.
This is similar to our ``continued training via gradient adaptation'' (Sec \ref{sec:consistency-via-continued-training}) setting, which we believe is a strong baseline to consider when studying meta-RL methods for OOD adaptation.

\section{Consistency}

This paper investigates how theoretical consistency translates into ``practical consistency'', i.e., the ability of a meta-RL algorithm to adapt to OOD tasks, given a large but limited adaptation budget. 
Below, we
define ways to quantify practical consistency (Sec \ref{sec:practical-consistency}), 
outline settings where consistency matters (Sec \ref{sec:practical-settings}), 
and propose a way to achieve practical consistency in context-based methods (Sec \ref{sec:consistency-via-continued-training}). 

\subsection{Quantifying Practical Consistency}
\label{sec:practical-consistency}

The definition of theoretical consistency (Sec \ref{subsec:theoretical-consistency}) in meta-RL assumes that policy gradient methods are consistent with arbitrary parameter initialization, which might not hold true in practice. 
To investigate whether meta-RL algorithms that are labelled as consistent have a practical advantage, we therefore propose two measures to quantify adaptation performance and speed empirically. 

\begin{definition}[Consistency Score]
    For an agent $\pi$ with initial weights $\theta_0$ and meta-trained weights $\theta_N$, 
    its consistency score $c_{\text{score}}$ for frame budget $F$ in an MDP $M$ is defined as
    \begin{equation}
        c_{\text{score}} 
        =
        \frac{
            R_{g(\pi_{\theta_N}, F, M)} - R_{\pi_{\theta_0}}
        }{
            R_{g(\pi_{\theta_0}, F, M)} - R_{\pi_{\theta_0}}
        }, 
    \end{equation}
    where $R_\pi$ is the return of a policy $\pi$ in the MDP $M$,
    and
    where $g(\pi, F, M):\Pi\rightarrow\Pi$ returns the policy $\pi$ after training for a given frame budget $F$ on $M$.
\end{definition}
The consistency score aims to measure: When adapting the meta-trained agent on the test task, what is the final performance, and how does this compare to training from scratch?

\begin{definition}[Consistency Rate]
    For an agent $\pi$ with initial weights $\theta_0$ and meta-trained weights $\theta_N$, 
    its consistency rate $c_{\text{rate}}\in\mathbb{R}_{\ge 0}$ for frame budget $F$ in an MDP $M$ is defined as
    \begin{equation}
        c_{\text{rate}}
        =
        \frac{t(\pi_{\theta_0}, F, M)}{t(\pi_{\theta_N}, F, M)},
    \end{equation}
    with $t(\pi, F, M):\Pi\rightarrow\mathbb{R}_{> 0}$, $t(\pi, F, M)=\arg\max R_\pi(f)$, where $R_\pi(f)$ is the return of policy $\pi$ at training frame $f$.
    In words, $t$ computes the number of frames it takes to train a policy $\pi$ to peak performance within the frame budget $F$.
\end{definition}
The consistency rate aims to measure: How long does it take the meta-trained agent to adapt compared to training from scratch?
Assuming that both policies reach the same maximum performance within the adaptation budget, then $c_{\text{rate}} > 1$ implies positive transfer, i.e., the meta-trained policy was able to adapt faster than training from scratch. Conversely, $c_{\text{rate}} < 1$ implies negative transfer. 
It is important to consider $c_{\text{rate}}$ of different algorithms together with their corresponding consistency score $c_{\text{score}}$, because if the policies do not converge to similar maximum performance within the adaptation budget, then $c_{\text{rate}}$ alone can be misleading. 

\subsection{Scenarios for Practical Consistency}
\label{sec:practical-settings}

We consider two scenarios where measuring consistency is useful in practice. 

\textbf{Single-task setting.}
Given a model meta-trained in $p_{\text{train}}$, we want to adapt it to a \emph{single} task in $p_{\text{test}}$.

\emph{Motivation.} 
The real world is noisy and constantly changing. 
Meta-learning promises flexible agents than can quickly adapt to such changes, but current methods are often specific to their training task distribution.
Overcoming this would have a positive impact for deploying agents in the real world.

\emph{Adaptation.} 
For all meta-RL algorithms, we first consider their default adaptation setting for learning on a single new task. 
For context-based methods, we also consider their consistent variants via gradient adaptation on the test task, as later introduced in Section \ref{sec:consistency-via-continued-training}. 

\emph{Evaluation.} 
We compare each meta-RL algorithm to an ``expert'', trained from scratch on each test task in $p_{\text{test}}$. 
The practical consistency metrics are computed w.r.t. this expert baseline. 

\textbf{Multi-task Setting.} 
Given a model meta-trained in $p_{\text{train}}$, we want to continue \emph{meta-learning} in $p_{\text{test}}$ to be able to quickly adapt to any task in that space. 

\emph{Motivation.} 
Imagine we have meta-learned a self-driving car that can adapt to different cities in one country, and want to further adapt to different cities in a new country.
As there are too many new tasks to solve in $p_{\text{test}}$, it may be more efficient to continue \emph{meta-training} in $p_{\text{test}}$ first, 
instead of adapting to each task separately.
Another example is when we first meta-learn in simulation (e.g., a factory robot that stacks things), then continue meta-learning in the real world to bridge the sim2real gap (in a training facility), before deploying the agent (in an actual factory).

\emph{Adaptation.} For all meta-RL algorithms, we first continue to meta-train the models on multiple tasks from $p_{\text{test}}$ following their default meta-training setup, then test the updated model on new tasks from $p_{\text{test}}$. 
All algorithms follow this same procedure, so we only have three methods to consider here. 
We call them MAML-CMT, RL$^2$-CMT and VariBAD-CMT to distinguish the continued multi-task meta-training (CMT) setting here from single-task gradient adaptation under the single-task setting. 

\emph{Evaluation.} 
We compare against a baseline which meta-learns in $p_{\text{test}}$ from scratch. 
The practical consistency values are computed w.r.t. this in-distribution meta-learning baseline. 

For both settings, it is not fair to directly compare the returns between different algorithms, as they each build on different components (e.g., network structure or the RL algorithm used) that can lead to varying absolute performance. 
The key is to compare different algorithms' practical consistency scores, which are computed w.r.t. their own baselines that use the same components. 

\subsection{Consistency of Context-based Methods via Gradient Adaptation}
\label{sec:consistency-via-continued-training}

As discussed in Section \ref{subsec:meta-rl-algos}, context-based methods are inconsistent under their default setting, when trying to adapt to a single OOD test task. 
We propose a simple gradient adaptation (GA) strategy to make these methods consistent at test time, i.e., instead of following the fixed meta-learned adaptation procedure, we continue training all agent components with gradient adaptation on the test task. 

\textbf{RL$^2$-GA.}
For RL$^2$, we do gradient updates to the entire RNN weights on the test task. The RNN is still unrolled and reset before each episode. 
In this way, the only change compared to the default adaptation setting of RL$^2$ is to update the whole model with gradient adaptation after each episode, 
but other choices are possible (e.g., freezing the hidden state at some point, or never resetting it).

\textbf{VariBAD-GA.}
For VariBAD, we update the VAE and the policy parameters via gradient updates. The RNN in the VAE encoder is still unrolled and reset before each episode.

In our experiments, we contrast these to the standard but inconsistent type of adaption that just unrolls the recurrent networks, i.e., RL$^2$-default and VariBAD-default.

\section{Experiments}

In this section we first introduce the tasks domains, and then present the results on the single-task and the multi-task scenario.
Detailed experiment settings and more results can be found in the appendix.

\subsection{Tasks}

\textbf{2D navigation.} The agent starts from the center of an unit circle, and has to navigate to an unknown goal position on the circle's edge. 
The reward is the negative distance between agent and goal.
We divide the circle into four goal spaces (left, right, up, bottom), resulting in $4\times4$ meta-train-test pairs. 

\textbf{2D sparse navigation.} A sparse variant of the above, where the reward is 0 if the agent's distance to the goal is larger than a threshold. 
We hypothesize that this can prove challenging when adapting to OOD tasks: 
the agent ought to explore outside the region where it has thus far seen goals. However, it might be overfit to that region and never get to see nonzero rewards from which it can learn.

\textbf{Cheetah velocity.} The goal is to control a half cheetah to run at a specific goal velocity. We consider 6 goal spaces ($[{-}3,{-}2]$, $[{-}2,{-}1]$, $[{-}1,0]$, $[0,1]$, $[1,2]$, $[2,3]$), 
resulting in $6\times6$ meta-train-test pairs. 
Negative velocities mean that the agent has to run backwards.
Intuitively, we expect that meta-training could help (e.g., forward $\rightarrow$ forward) or hurt (e.g., forward $\rightarrow$ backward) for OOD adaptation. 

\textbf{Cheetah velocity $\rightarrow$ cheetah direction.} The goal of cheetah direction is to control a half cheetah to run either forward or backward as far as possible. 
By looking at transfer from cheetah velocity to this task, we can investigate adaptation to tasks with a different reward structure.

\textbf{Cheetah direction $\leftrightarrow$ ant direction.} This setting is the hardest, as the two task spaces significantly differ in state space, action space and transition function. We expect there to be minimal positive transfer in this environment, or possibly negative transfer.
Ant has a larger state dimensionality, so we pad the observations in Cheetah with zeros. Ant also has a larger action dimension, so we extend the action dimension of cheetah to match that of ant, and only consider the action dimensions with practical meanings when computing the action loss for cheetah. 
\subsection{Results: Single-Task Setting}
\label{subsec:result-single-task}

\begin{table}[h]
\small
\begin{center}
\caption{Average consistency scores of different methods under the single-task setting. The score is averaged over all OOD meta-train-test pairs for each method.}
\label{tab:single-task-summary-score}
\begin{tabular}{l|c|c|c|c|c}
\toprule
Algorithm & Navi & Sparse-navi & Cheetah-vel & Cheetah-vel $\rightarrow$ dir & Cheeetah-dir $\leftrightarrow$ ant-dir \\
\midrule
MAML & 0.80 & 0.00 & \textbf{0.99} & \textbf{0.92} & \textbf{0.91} \\
\hline
RL$^2$-default & -0.49 & 0.05 & 0.00 & 0.22 & 0.00 \\
RL$^2$-GA & 0.52 & \textbf{0.42} & 0.78 & 0.80 & 0.39 \\
\hline
VariBAD-default & -0.20 & 0.03 & -0.05 & 0.27 & 0.09 \\
VariBAD-GA & \textbf{0.91} & 0.29 & \textbf{0.99} & \textbf{0.93} & 0.66 \\
\bottomrule
\end{tabular}
\end{center}
\end{table}

Table \ref{tab:single-task-summary-score} shows the average consistency scores of different algorithms under the single-task setting. 
We summarize our main findings from different tasks under the single-task setting as follows.

\begin{figure}[t]
    \centering
    \includegraphics[width=\textwidth]{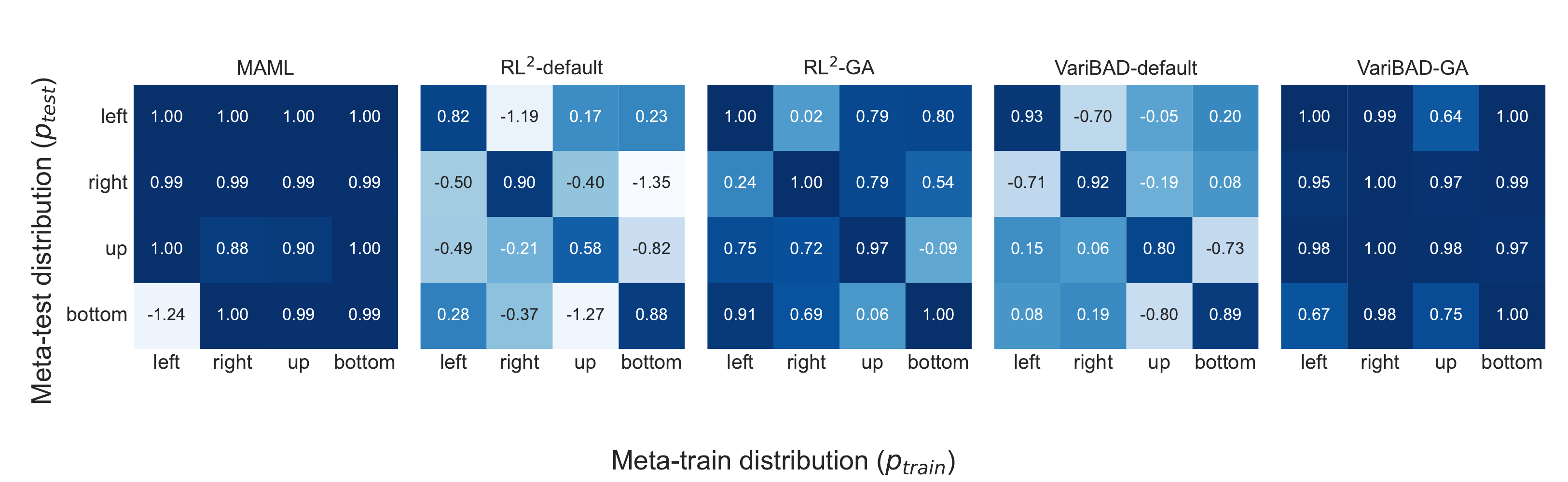}
    \caption{Consistency scores of different meta-RL methods on 2D navigation, in the single-task setting.
    Darker color means better asymptotic performance. 
    We suggest to look at consistency scores (Fig \ref{fig:navigation-consistency-score-single-task}) and consistency rates (Fig \ref{fig:navigation-rate-single-task}) together, as in some cases an agent quickly converges to a sub-optimal solution (high consistency rate, low consistency score), which is not desirable in practice. 
    }
    \label{fig:navigation-consistency-score-single-task}
\end{figure}

\begin{figure}[t]
    \centering
    \includegraphics[width=\textwidth]{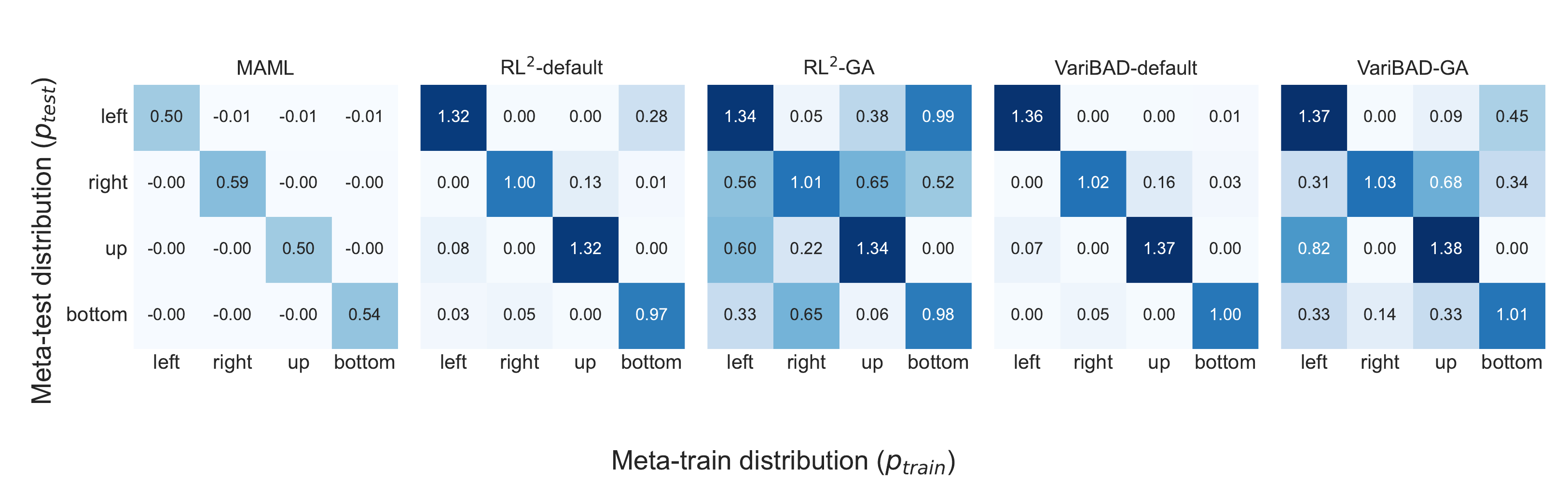}
    \caption{Consistency scores of different meta-RL algorithms on 2D sparse navigation, in the single-task setting. 
    }
    \label{fig:sparse-navigation-consistency-score-single-task}
\end{figure}

\begin{enumerate}[leftmargin=*,topsep=0pt]
\item The theoretically consistent MAML indeed adapts OOD quite well on most task domains except for sparse navigation. 
MAML fails on sparse navigation as it only explores within a limited area in the meta-trained direction, and gets no training signal. 

\item Context-based methods have low consistency scores, i.e., poor asymptotic performance, under their default setting. 
Gradient adaptation significantly improves this score, indicating that we can indeed make context-based methods consistent via gradient adaptation. Furthermore, VariBAD-GA achieves performance similar to or even better than the theoretically consistent MAML.

\item Aligned with our intuition, the distribution shifts between $p_{\text{train}}$ and $p_{\text{test}}$ are correlated with the consistency score and rate, i.e., the smaller the distribution shift, the better and faster a model adapts to OOD tasks. For example, on cheetah velocity, cells closer to the diagonal line usually have higher values (Figure \ref{fig:halfcheetah-vel-consistency-score-single-task}). On 2D navigation and sparse navigation, transfer from the opposite direction generally has lower scores than transferring from a neighboring space (Figures \ref{fig:navigation-consistency-score-single-task} and \ref{fig:sparse-navigation-consistency-score-single-task}). 

\item When adapting context-based methods via gradient adaptation, VariBAD-GA performs better than RL$^2$-GA on most tasks. 
A possible reason is that VariBAD divides the labour between task encoder and policy, which makes learning easier \citep{zintgraf2019varibad}, and might also help OOD adaptation. 
\end{enumerate}

We discuss the results of the navigation tasks below. 
See Appendix \ref{appendix:single-task-results} for full results on other tasks.

\begin{figure}[t]
    \centering
    \includegraphics[width=0.8\textwidth]{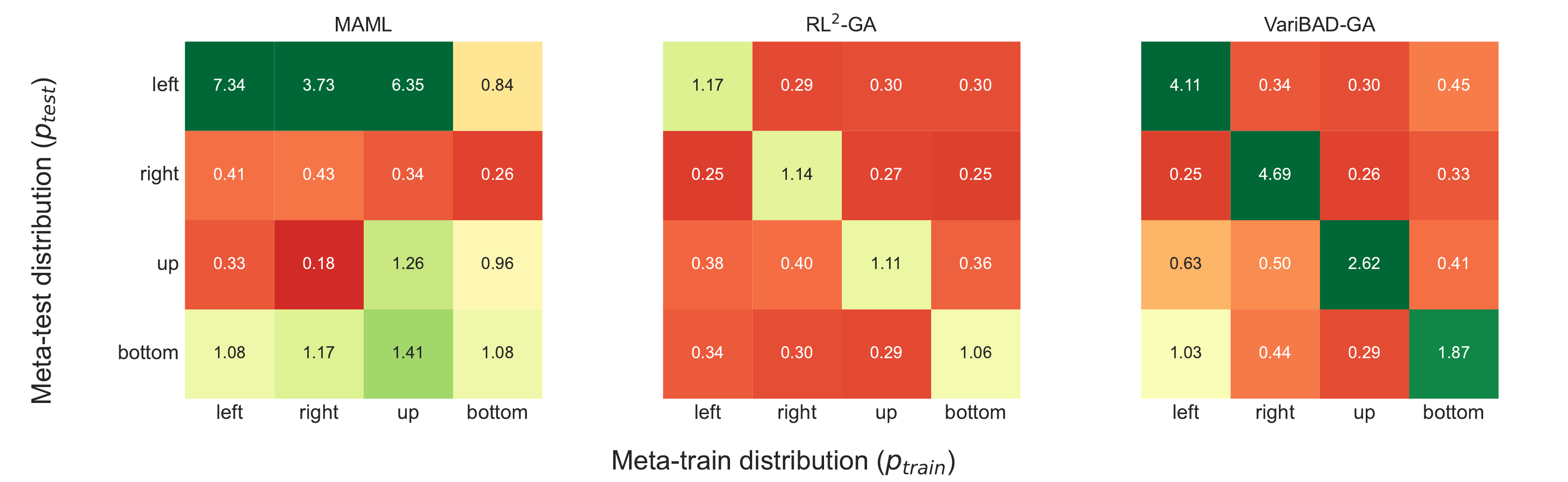}
    \caption{Consistency rates of different meta-RL algorithms on 2D navigation, in the single-task setting. 
    {\color{ForestGreen}Green} represents positive transfer, while {\color{Red}red} represents negative transfer. 
    We omit RL$^2$-default and VariBAD-default, as only a few episodes is required for fast adaptation under their default setting, which is much smaller than the adaptation budgets of other methods. 
    }
    \label{fig:navigation-rate-single-task}
\end{figure}

\textbf{2D navigation with dense rewards.} 
The consistency score and rate are shown in Figures \ref{fig:navigation-consistency-score-single-task} and \ref{fig:navigation-rate-single-task}. 
Main results: 
(1) For MAML, most meta-train-test pairs achieve consistency score close to 1, which means OOD adaptation with MAML performs well in general. 
(2) Context-based methods cannot adapt OOD under their default setting, as illustrated by the low consistency scores in the off-diagonal cells of RL$^2$-default and VariBAD-default. 
(3) Gradient adaptation significantly improves the consistency scores of context-based methods, especially for VariBAD. 
(4) Context-based methods get consistency rate smaller than 1 in most OOD tasks, which indicates negative transfer. MAML is more efficient at OOD adaptation (with both positive and negative transfer). 

\begin{figure}[t]
    \centering
    \includegraphics[width=0.8\textwidth]{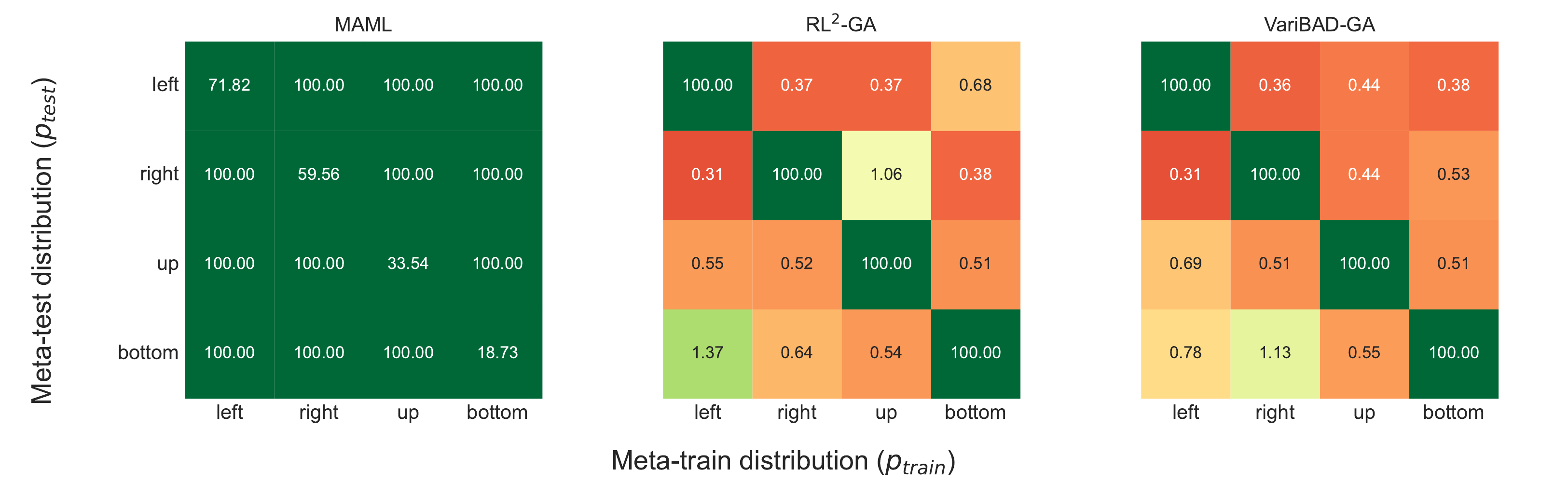}
    \caption{Consistency rates of different meta-RL algorithms on 2D sparse navigation in the single-task setting. 
    MAML has very high consistency rates here because it quickly converges to a poor solution. 
    The maximal rate is 100 since we add a constant of 0.01 to the denominator to avoid division by 0. 
    }
    \label{fig:sparse-navigation-rate-single-task}
\end{figure}

\textbf{2D navigation with sparse rewards.} 
The consistency score and rate are shown in Figures \ref{fig:sparse-navigation-consistency-score-single-task} and \ref{fig:sparse-navigation-rate-single-task}. 
Main results: 
(1) In contrast to its consistent performance in the dense-reward setting, MAML fails for OOD adaptation on this sparse navigation task. We believe this is due to poor exploration in regions beyond the meta-training tasks, leading to a lack of learning signal (all rewards are zero).
(2) Context-based methods fail under their default setting, but partly solve this task via gradient adaptation. 
We attribute their advantages over MAML on this task to better exploration introduced by some specific components in their implementations. 
We discuss this in more detail in Appendix \ref{appendix:sparse-navigation-analysis}, where we also visualise the policy's trajectories that show how the meta-learned prior indeed hinders exploration. 
See also a related discussion on universality by \cite{finn2018learning}.


\subsection{Results: Multi-Task Setting}
\label{subsec:result-multi-task}

\begin{table}[h]
\small
\begin{center}
\caption{Average consistency scores of different methods under the multi-task setting. The score is averaged over all OOD meta-train-test pairs for each method.}
\label{tab:multi-task-summary-score}
\begin{tabular}{l|c|c|c|c|c}
\toprule
Algorithm & Navi & Sparse-navi & Cheetah-vel & Cheetah-vel $\rightarrow$ dir & Cheeetah-dir $\leftrightarrow$ ant-dir \\
\midrule
MAML-CMT & 0.70 & -0.02 & 0.80 & 0.73 & 0.59 \\
\hline
RL$^2$-CMT & 0.84 & 0.71 & 0.98 & 1.04 & 0.35 \\
\hline
VariBAD-CMT & \textbf{0.99} & \textbf{0.79} & \textbf{1.05} & \textbf{1.06} & \textbf{0.66} \\
\bottomrule
\end{tabular}
\end{center}
\end{table}

Table \ref{tab:multi-task-summary-score} illustrates the average consistency scores of different algorithms under the multi-task setting. 
The general trends under the multi-task setting are consistent with what we have observed under the single-task setting, so we refer the reader to the detailed results on each task in Appendix \ref{appendix:multi-task-results}. 

One key difference to the single-task setting is that we observe a noticeable increase in the summary score of RL$^2$ and VariBAD on most tasks. 
The reason why context-based methods perform better under the multi-task setting may be that the context encoder is trained with more diverse trajectories from different underlying dynamics and thus can learn to better infer the context compared to overfitting to a single environment under the single-task setting. 
\section{Conclusion and Future Work}

\paragraph{Conclusion.}
In this work, we empirically studied how well different meta-RL algorithms adapt to OOD tasks. We summarize our main findings as follows:
\begin{enumerate}
    \item Consistent with their theoretical properties, gradient-based methods can adapt to OOD tasks, while context-based methods can't. However, context-based methods adapt as well or better to OOD tasks by simply continuing to train the meta-learned policy via gradient adaptation. 
    \item Theoretical consistency of meta-RL algorithms may not hold empirically. We identify one such failure case, where the agent explores only in a restricted region due to the meta-learned inductive bias and thus fails to reach OOD goals in navigation tasks with sparse rewards. 
    \item On most tasks, we observe a correlation between the extent of task distribution shift and adaptation efficiency, i.e., the smaller the distribution shift, the faster the agent adapts in the OOD task space. However, we also observe positive transfer between tasks which are intuitively different (e.g., Cheetah and Ant). 
    This implies a need for accurate quantification of distribution shift, such as those in the transfer learning literature \citep{ammar2014automated}. 
\end{enumerate}

In summary, we believe that theoretical consistency is indeed a desirable property, and inconsistent algorithms can easily be made consistent. 
Even with the simple strategy of updating the entire meta-trained agent, the originally inconsistent context-based methods were able to adapt OOD. 

\paragraph{Future Work.}
A drawback of the ``gradient adaptation'' approach is that it is necessary to explicitly start this process (i.e., know that we are out of distribution). 
An alternative is to always keep updating everything regardless of whether we are OOD or not. In our experiments however (see Appendix \ref{appendix:adaptation-curve}), we found that with our current setup, this can lead to a slight dip in performance when the test tasks are in-distribution. Future work could focus on how to tackle this instability for in-distribution adaptation or how to automatically detect distribution shifts for adaptation strategy selection. 

We also found that insufficient exploration can, unsurprisingly, lead to failure on OOD tasks.
Another potential avenue of future work is therefore to develop agents that, when faced with an OOD task, start exploring more widely again.

In this paper, we did not consider meta-RL algorithms that meta-learn aspects other than a policy, such as a loss function, how to set hyperparameters, or training environments. 
In these methods, inconsistency might be harder to overcome than with context-based methods.

A remaining open question is where the differences in OOD adaptation capabilities of meta-RL algorithms stem from. 
We conclude this paper by listing in Table \ref{tab:discuss} some hypotheses about the potential factors that may have positive or negative effects on the learning process.
We leave it for future work to develop a better understanding of how exploration, optimization and other issues influence OOD adaptation through model initialization. 
\begin{table}[h]
\begin{center}
\caption{Hypotheses about the potential positive and negative effects of using a meta-learned model to initialize the learning process on OOD tasks compared to random initialization.}
\label{tab:discuss}
\begin{tabular}{l|p{5.4cm}|p{5.4cm}}
\toprule
Issue & Positive effects & Negative effects \\
\midrule
Exploration & 
The initial policy saves unnecessary exploration & 
The initial policy hinders exploration in the right direction \\
\hline
Optimization & 
Easier to optimize from the initial policy in the parameter space, e.g., fewer local minima in the loss function surface on the path from the initial policy to the solution policy, shorter distance to the solution parameters & 
Harder to optimize from the initial policy in the parameter space, e.g., the loss function surface of the test task is flat around the initial parameters, greater distance to the solution parameters \\
\bottomrule
\end{tabular}
\end{center}
\end{table}

\section*{Acknowledgments and Disclosure of Funding}
Zheng Xiong is supported by UK EPSRC CDT in Autonomous Intelligent Machines and Systems (grant number EP/S024050/1) and AWS.
Luisa Zintgraf is supported by the 2017 Microsoft Research PhD Scholarship Program, and the 2020 Microsoft Research EMEA PhD Award.
Jacob Beck is supported by the Oxford-Google DeepMind Doctoral Scholarship.
Risto Vuorio is supported by EPSRC Doctoral Training Partnership Scholarship and Department of Computer Science Scholarship. 
This work was supported by a generous equipment grant and a donated {DGX-1} from NVIDIA.

\bibliography{reference}
\bibliographystyle{apalike}

\newpage
\appendix
\section{Experimental Setup}
\label{appendix:experiment-setup}

The episode length is set to 100 for navigation, and 200 for Mujoco. 
For all experiments, results are averaged over 3 seeds. 
The learning budgets are shown in Table \ref{tab:budget}. 
For MAML, we set the batch size of both outer and inner loop to 10, tune the learning rate of inner loop in $\{ 0.1, 0.05, 0.02, 0.01 \}$, and follow the default setting in the PyTorch implementation of MAML\footnote{\url{https://github.com/tristandeleu/pytorch-maml-rl}} for other configurations. 
For RL$^2$ and VariBAD, we follow the default setting in the official code of VariBAD\footnote{\url{https://github.com/lmzintgraf/varibad}}. 
As the meta-training process is unstable in some cases (especially for RL$^2$), we use the model with the highest evaluation returns during meta-training, instead of the final model after meta-training, for adaptation. 

\begin{table}[h]
\begin{center}
\caption{Frame number for each task and different learning stages.}
\label{tab:budget}
\begin{tabular}{c|c|c|c}
\toprule
Task & Meta-train & Single-task adaptation & Multi-task adaptation \\
\midrule
Navigation & \makecell[c]{10M (MAML)\\5M (RL$^2$, VariBAD)} & 2M & \makecell[c]{10M (MAML)\\5M (RL$^2$, VariBAD)} \\
\hline
Sparse navigation & 20M & 5M & 20M \\
\hline
Cheetah velocity & 10M & 2M & 10M \\
\hline
Cheetah vel $\rightarrow$ dir & 10M & 5M & 10M \\
\hline
Cheetah dir $\leftrightarrow$ ant dir & 20M & 10M & 20M \\
\bottomrule
\end{tabular}
\end{center}
\end{table}

Empirically, we compute the consistency score by first smoothing the learning curve with a sliding window of length 5, then using the highest return of the smoothed learning curve as $R_\pi$. 
For the consistency rate, we compute $t(\pi, F, M)$ as the frame number the agent needs to reach a return of $0.98 R_\pi$ on the smoothed learning curve (as the maximal return $R_\pi$ may be achieved long after convergence due to small fluctuation in the learning curve). 

\section{Further Experimental Results}

\subsection{Single-Task Setting}
\label{appendix:single-task-results}

\begin{figure}[t]
    \centering
    \includegraphics[width=\textwidth]{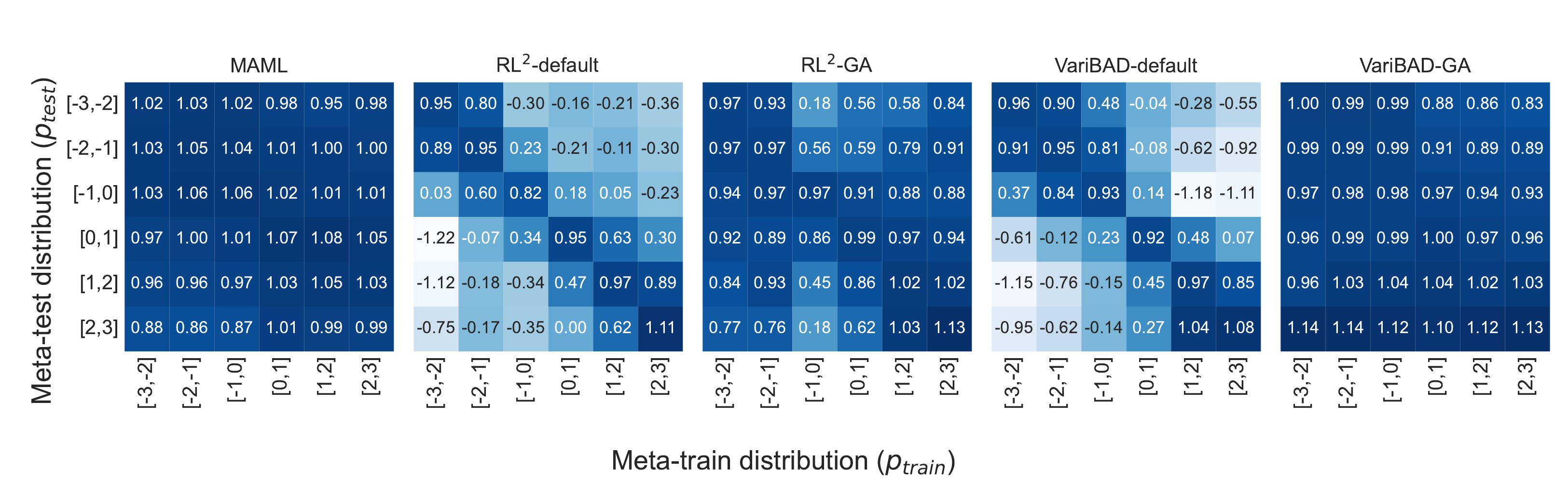}
    \caption{Consistency scores of different meta-RL methods on cheetah velocity; single-task setting. 
    }
    \label{fig:halfcheetah-vel-consistency-score-single-task}
\end{figure}

\begin{figure}[t]
    \centering
    \includegraphics[width=\textwidth]{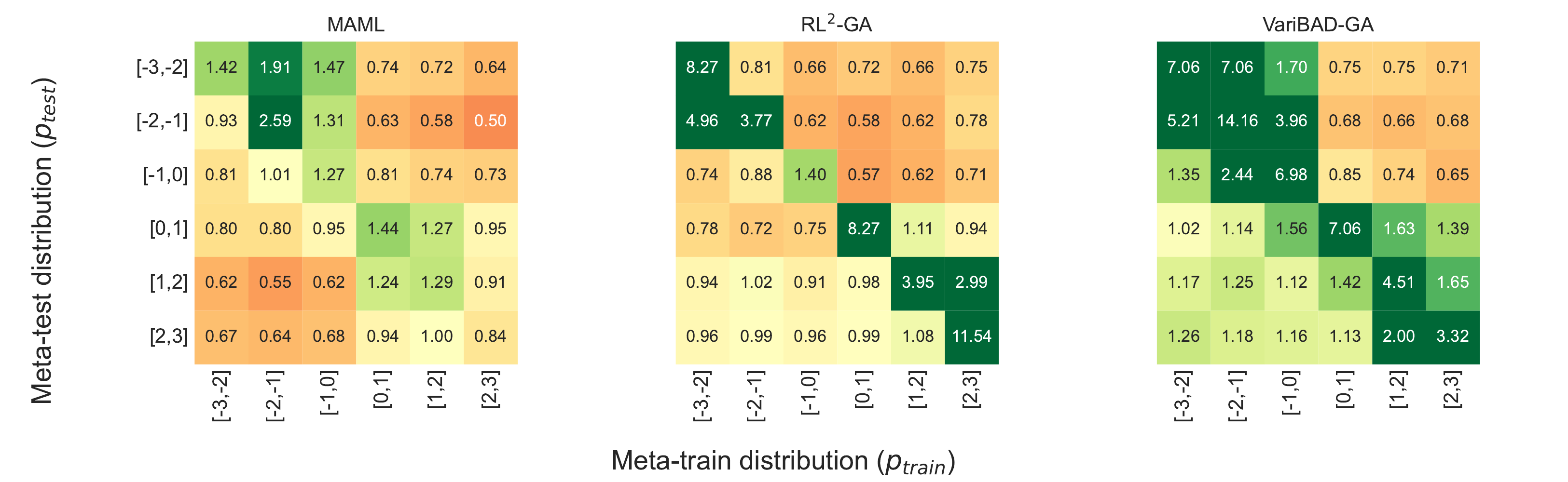}
    \caption{Consistency rates of different meta-RL methods on cheetah velocity; single-task setting. 
    }
    \label{fig:halfcheetah-vel-rate-single-task}
\end{figure}

\textbf{Cheetah velocity.} 
The consistency score and rate are shown in Figures \ref{fig:halfcheetah-vel-consistency-score-single-task} and \ref{fig:halfcheetah-vel-rate-single-task}. 
Main results:
(1) Under their default setting, context-based methods can only partly solve OOD tasks with small shifts in goal velocity, and their performance significantly improves via gradient adaptation. 
(2) MAML, RL$^2$-GA and VariBAD-GA show positive OOD adaptation, i.e., similar convergence performance and better learning efficiency compared to the ``expert'' baseline, if the distribution shift is not too large. Specifically, if we use the median value of each velocity interval to represent the corresponding task space, and use 0 to represent the ``expert'', then there is a general trend that OOD adaptation is more efficient than single-task learning from scratch if $|b-a|<|b-0|$, where $a$ and $b$ are the numbers representing the meta-training and meta-test space respectively. 
(3) For the context-based methods via gradient adaptation, VariBAD generally outperforms RL$^2$ w.r.t. both adaptation efficiency and final performance, especially when the distribution shift is large. 

\begin{figure}[t]
    \centering
    \includegraphics[width=\textwidth]{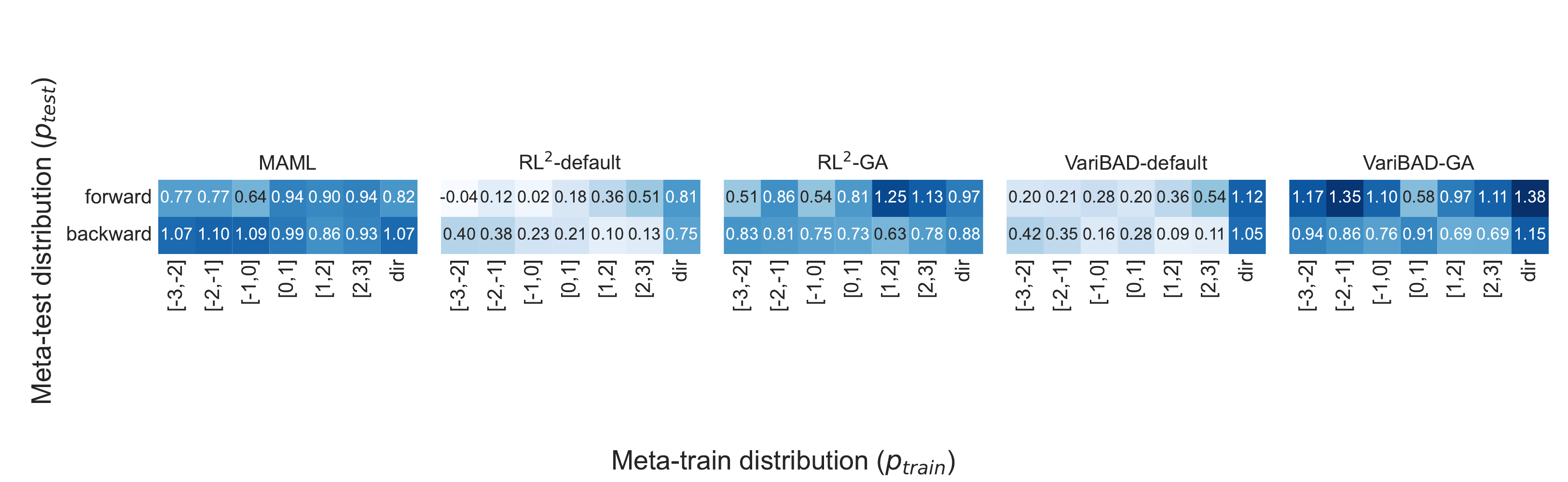}
    \caption{Consistency scores of different meta-RL algorithms on cheetah direction; single-task setting. 
    }
    \label{fig:cheetah-dir-consistency-score-single-task}
\end{figure}

\begin{figure}[t]
    \centering
    \includegraphics[width=\textwidth]{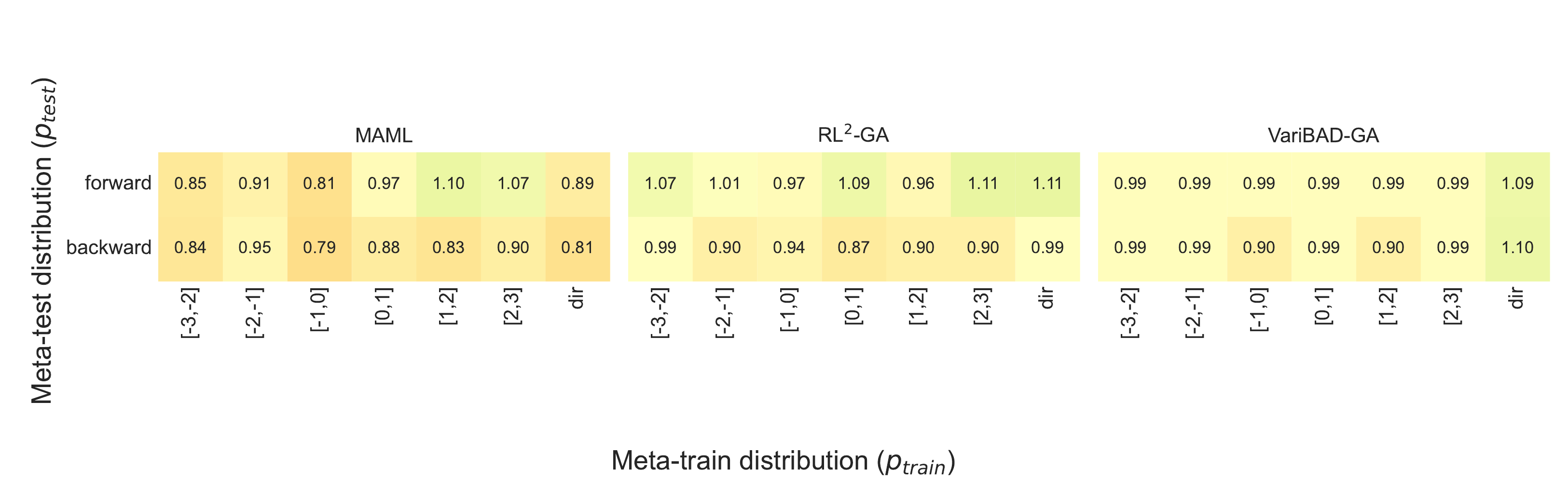}
    \caption{Consistency rates of different meta-RL algorithms on cheetah direction; single-task setting. 
    }
    \label{fig:cheetah-dir-rate-single-task}
\end{figure}

\textbf{Cheetah velocity $\rightarrow$ cheetah direction.} 
The consistency score and rate are shown in Figures \ref{fig:cheetah-dir-consistency-score-single-task} and \ref{fig:cheetah-dir-rate-single-task}. 
Main results: 
(1) Consistent with the results on other tasks, gradient adaptation significantly improves the performance of context-based methods. 
(2) Surprisingly, an agent trained to run at a specific velocity in one direction can transfer quite well to the task of running as far as possible in the opposite direction. One possible reason might be that although the goal direction is opposite, the agent can still learn some useful prior knowledge from meta-training, such as how to run (fast), which is essential for solving the test task. 

\begin{figure}[t]
    \centering
    \includegraphics[width=\textwidth]{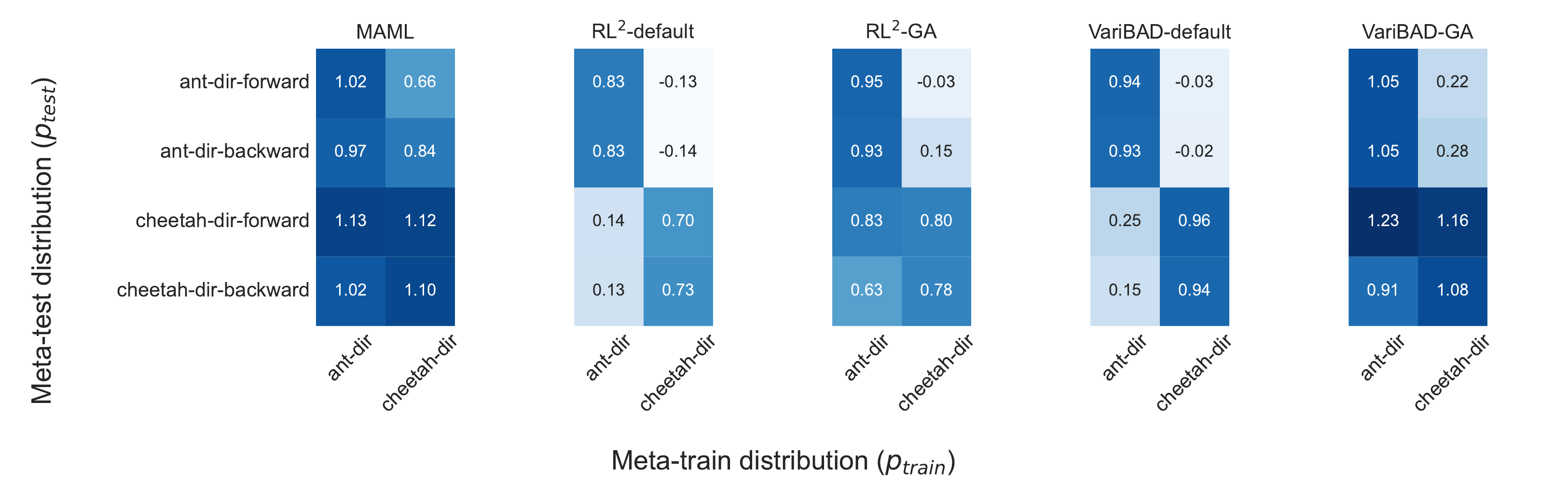}
    \caption{Consistency scores of different meta-RL algorithms on cheetah direction $\leftrightarrow$ ant direction; single-task setting. 
    }
    \label{fig:ant-cheetah-dir-consistency-score-single-task}
\end{figure}

\begin{figure}[t]
    \centering
    \includegraphics[width=\textwidth]{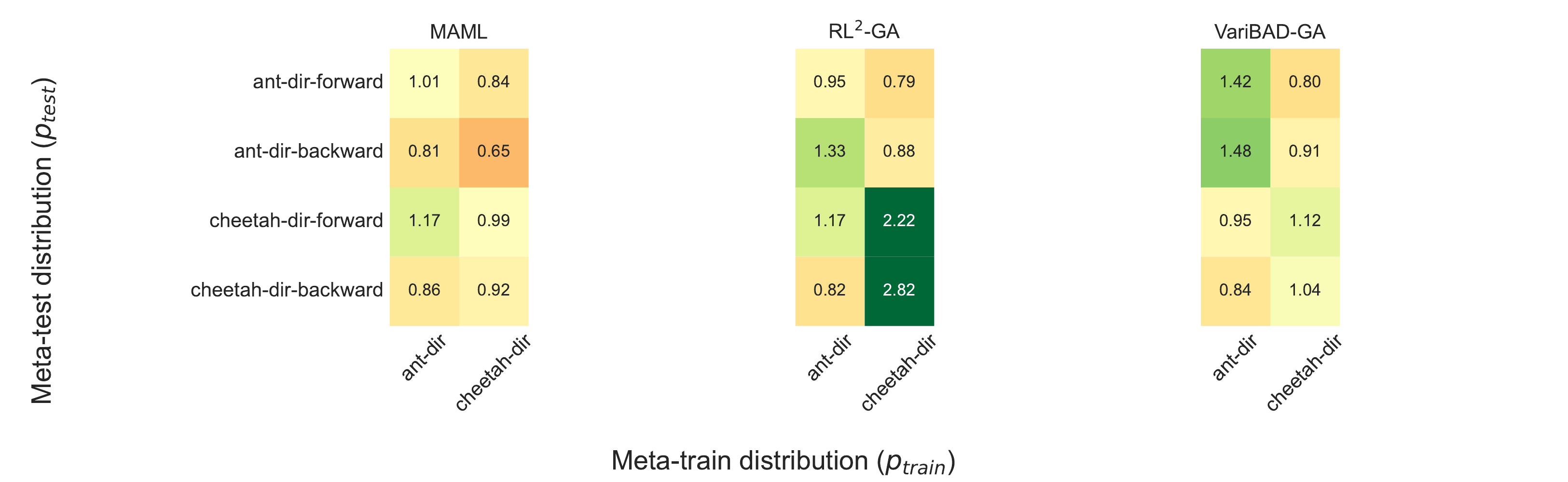}
    \caption{Consistency rates of different meta-RL algorithms on cheetah-direction $\leftrightarrow$ ant-direction; single-task setting. 
    }
    \label{fig:ant-cheetah-dir-rate-single-task}
\end{figure}

\textbf{Cheetah direction $\leftrightarrow$ ant direction.}
The consistency score and rate are shown in Figures \ref{fig:ant-cheetah-dir-consistency-score-single-task} and \ref{fig:ant-cheetah-dir-rate-single-task}. 
Main results: 
(1) Gradient adaptation significantly improves the performance of context-based methods. 
(2) The transfer between the two tasks is asymmetric, i.e., ant direction $\rightarrow$ cheetah direction performs much better than cheetah direction $\rightarrow$ ant direction.
The main reason may be that the agent observes additional inputs and has more action outputs when transferring from cheetah to ant.

\subsection{Multi-Task Setting}
\label{appendix:multi-task-results}

\begin{figure}[t]
    \centering
    \includegraphics[width=0.8\textwidth]{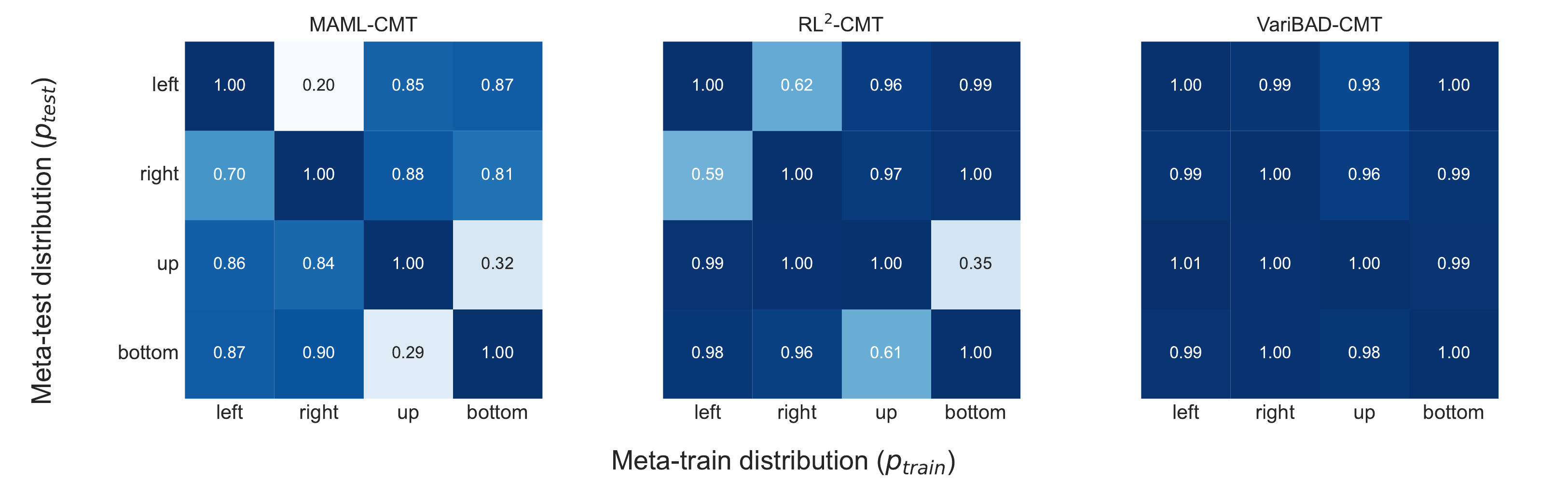}
    \caption{Consistency scores of different meta-RL algorithms on 2D navigation; multi-task setting. 
    }
    \label{fig:navigation-consistency-score}
\end{figure}

\begin{figure}[t]
    \centering
    \includegraphics[width=0.8\textwidth]{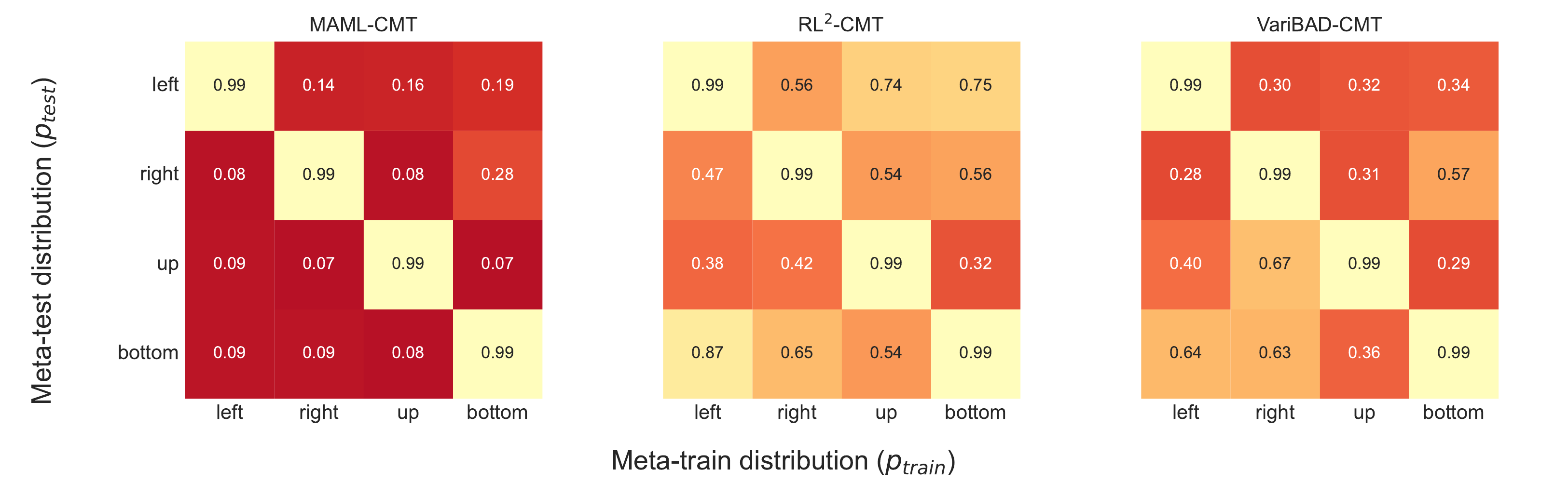}
    \caption{Consistency rates of different meta-RL algorithms on 2D navigation; multi-task setting. 
    }
    \label{fig:navigation-rate}
\end{figure}

\textbf{2D navigation with dense rewards.} 
The consistency score and rate are shown in Figures \ref{fig:navigation-consistency-score} and \ref{fig:navigation-rate}. 
Main results: 
(1) W.r.t. asymptotic performance, VariBAD achieves OOD adaptation performance very similar to in-distribution learning, while MAML and RL$^2$ are worse when the meta-test goal direction is opposite to that of meta-training. 
(2) W.r.t. learning efficiency, none of the algorithms can adapt to the OOD space faster than in-distribution meta-learning, as shown by the fact that no off-diagonal cell has consistency rate larger than 1. 

\begin{figure}[t]
    \centering
    \includegraphics[width=0.8\textwidth]{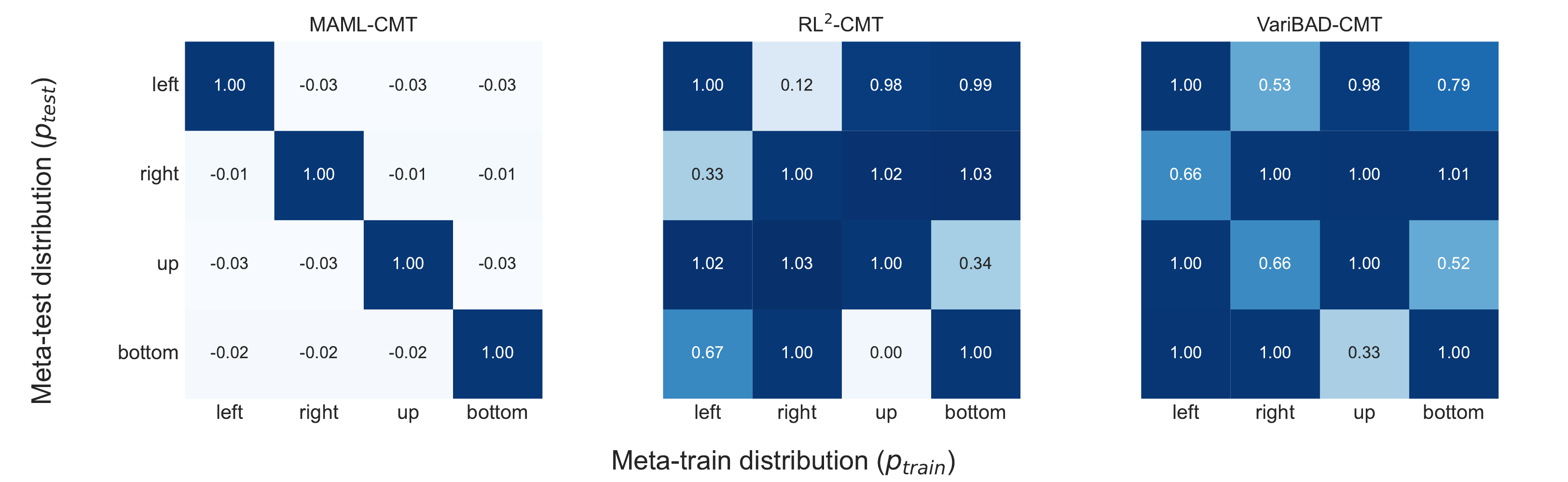}
    \caption{Consistency scores of different meta-RL algorithms on 2D sparse navigation; multi-task setting. 
    }
    \label{fig:sparse-navigation-consistency-score}
\end{figure}

\begin{figure}[t]
    \centering
    \includegraphics[width=0.8\textwidth]{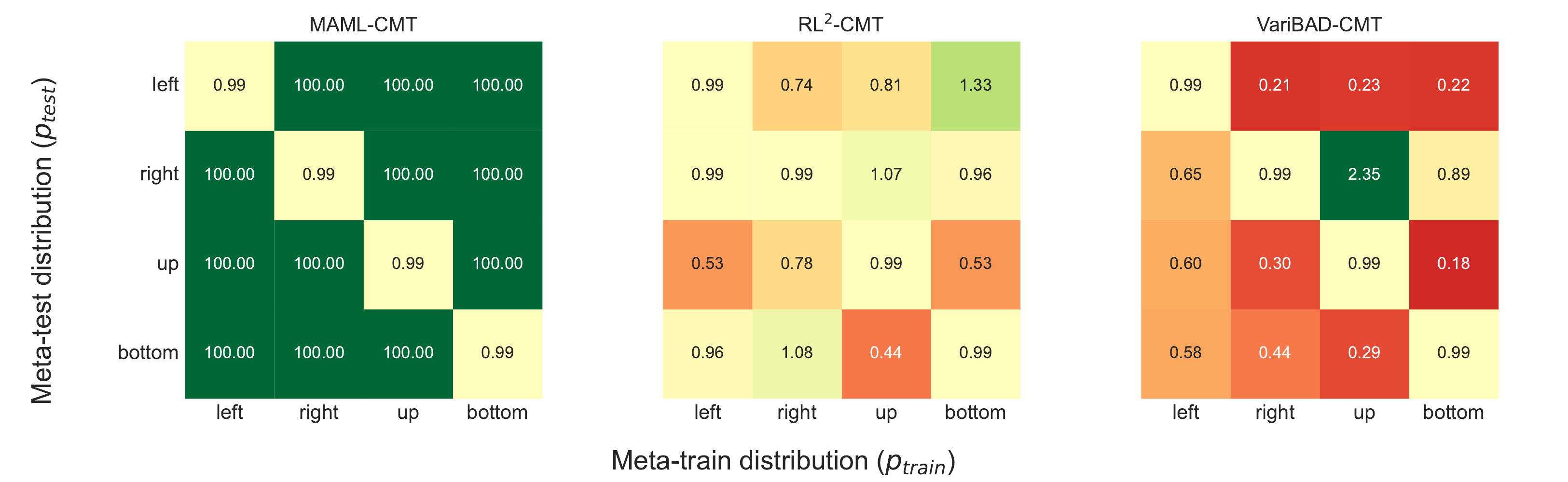}
    \caption{Consistency rates of different meta-RL algorithms on 2D sparse navigation; multi-task setting. 
    }
    \label{fig:sparse-navigation-rate}
\end{figure}

\textbf{2D navigation with sparse rewards.} 
The consistency score and rate are shown in Figures \ref{fig:sparse-navigation-consistency-score} and \ref{fig:sparse-navigation-rate}. 
Main results: 
(1) MAML totally fails with OOD adaptation due to poor exploration, similar to its result on the same task under the single-task setting. 
(2) RL$^2$ and VariBAD in some cases can solve this challenging OOD adaptation task with sparse rewards, while the task also becomes much harder for them if the model is transferred to the opposite direction.

\begin{figure}[t]
    \centering
    \includegraphics[width=\textwidth]{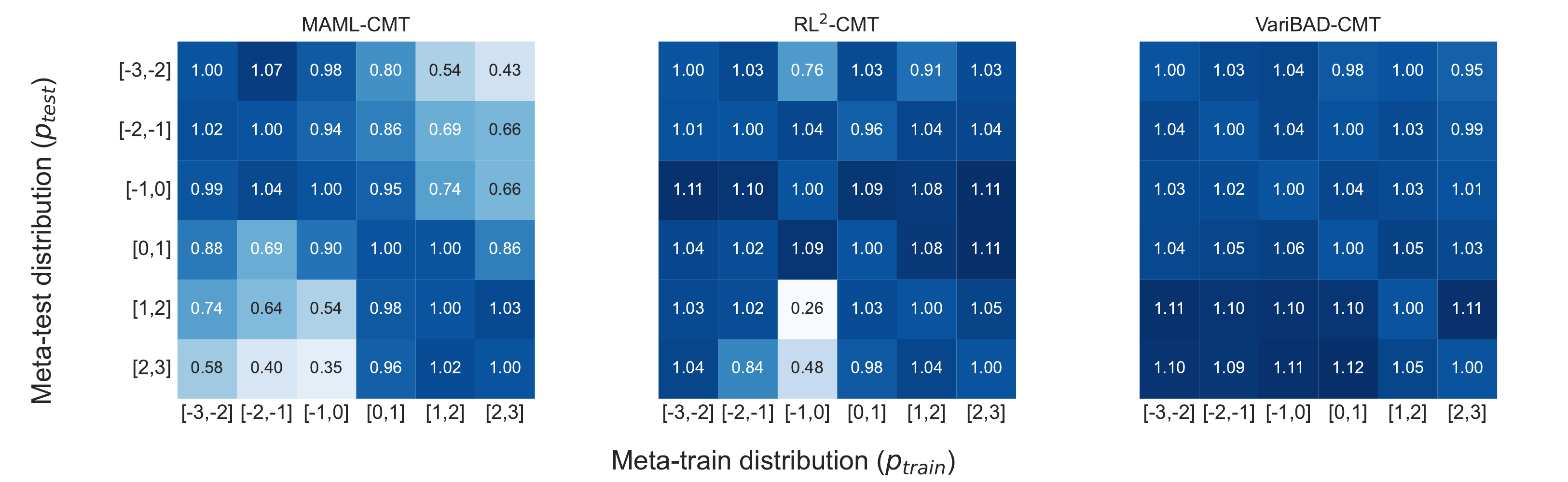}
    \caption{Consistency scores of different meta-RL algorithms on cheetah velocity; multi-task setting. 
    }
    \label{fig:halfcheetah-vel-consistency-score}
\end{figure}

\begin{figure}[t]
    \centering
    \includegraphics[width=\textwidth]{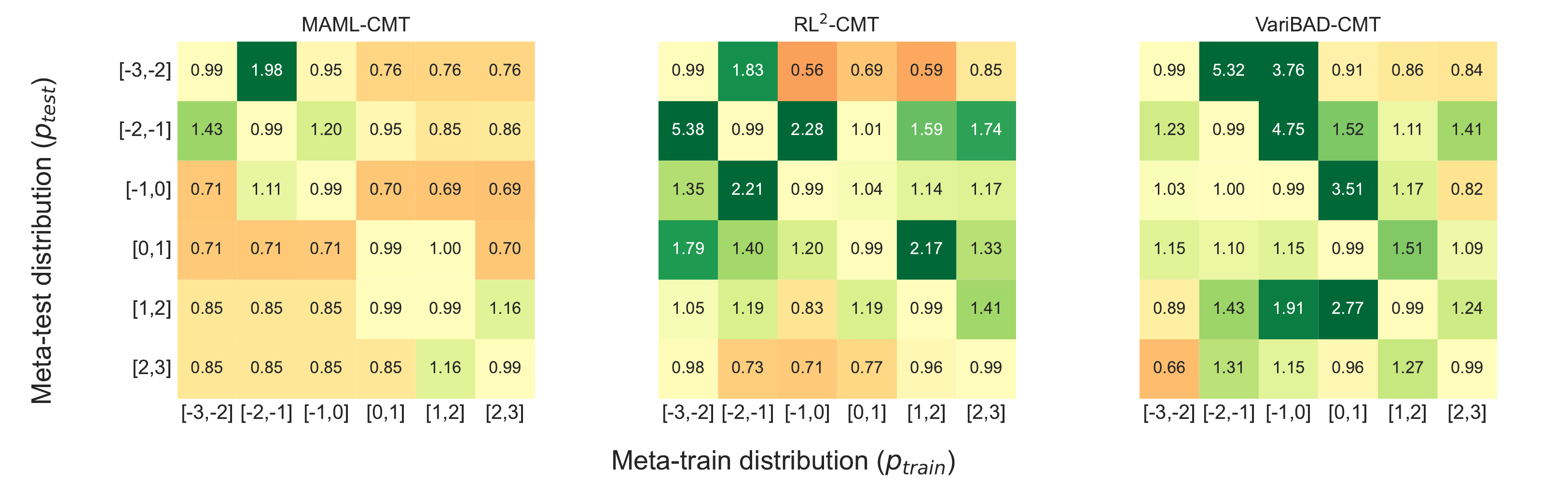}
    \caption{Consistency rates of different meta-RL algorithms on cheetah velocity; multi-task setting. 
    }
    \label{fig:halfcheetah-vel-rate}
\end{figure}

\textbf{Cheetah velocity.} 
The consistency score and rate are shown in Figures \ref{fig:halfcheetah-vel-consistency-score} and \ref{fig:halfcheetah-vel-rate}. 
Main results: 
(1) W.r.t. consistency score, MAML's performance drops as the distribution shift increases, while RL$^2$ and VariBAD are generally more robust (expect for several meta-train-test pairs of RL$^2$). 
(2) W.r.t. consistency rate, generally all methods adapt to OOD tasks more efficiently than in-distribution learning if the distribution shift is small, while RL$^2$ and VariBAD better maintain their high adaptation efficiency as distribution shift increases compared to MAML. 

\begin{figure}[t]
    \centering
    \includegraphics[width=\textwidth]{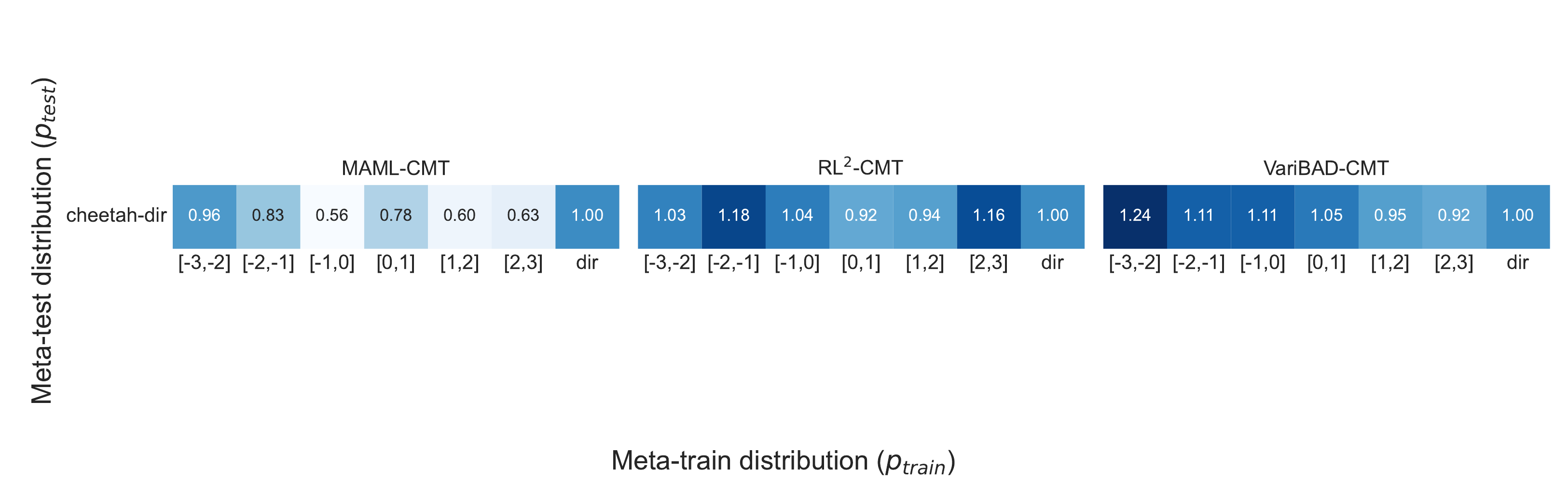}
    \caption{Consistency scores of different meta-RL algorithms on cheetah direction; multi-task setting. 
    }
    \label{fig:cheetah-dir-consistency-score}
\end{figure}

\begin{figure}[t]
    \centering
    \includegraphics[width=\textwidth]{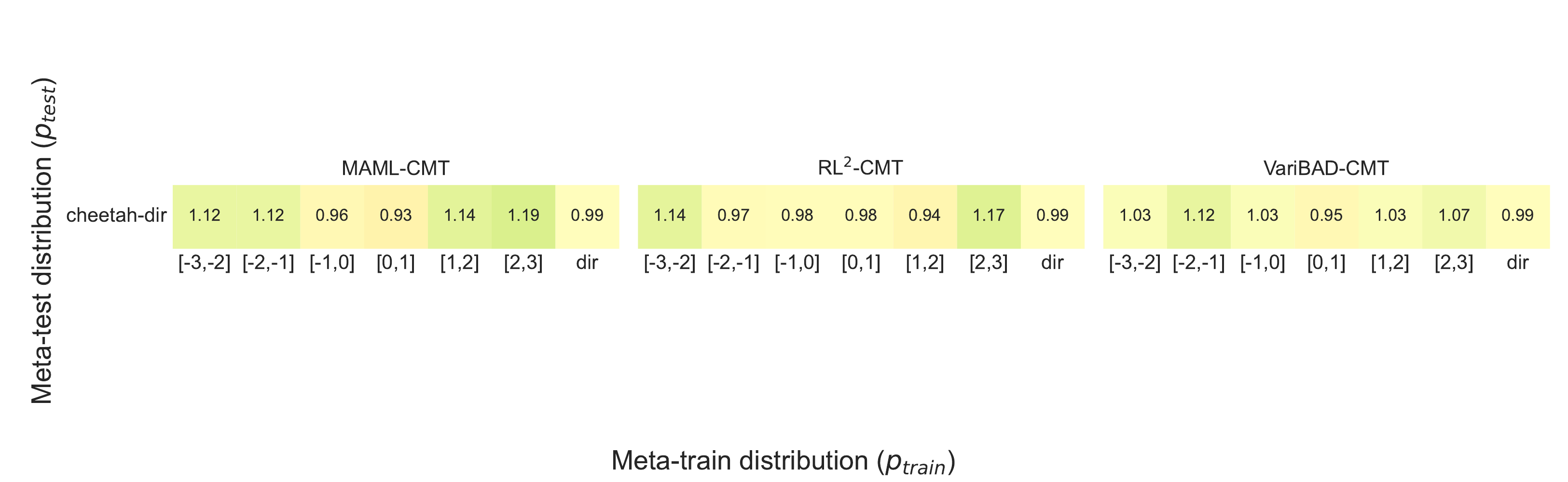}
    \caption{Consistency rates of different meta-RL algorithms on cheetah direction; multi-task setting. 
    }
    \label{fig:cheetah-dir-rate}
\end{figure}

\textbf{Cheetah velocity $\rightarrow$ cheetah direction.} 
The consistency score and rate are shown in Figures \ref{fig:cheetah-dir-consistency-score} and \ref{fig:cheetah-dir-rate}. 
Main results: 
(1) For MAML, the OOD learning performance is slightly lower than in-distribution learning. 
(2) For RL$^2$ and VariBAD, although the meta-trained models are only capable of running in one direction within a specific velocity range, they adapt quite well to the new task of running in both directions, and achieve similar learning efficiency compared to in-distribution meta-learning from scratch. 

\begin{figure}[t]
    \centering
    \includegraphics[width=\textwidth]{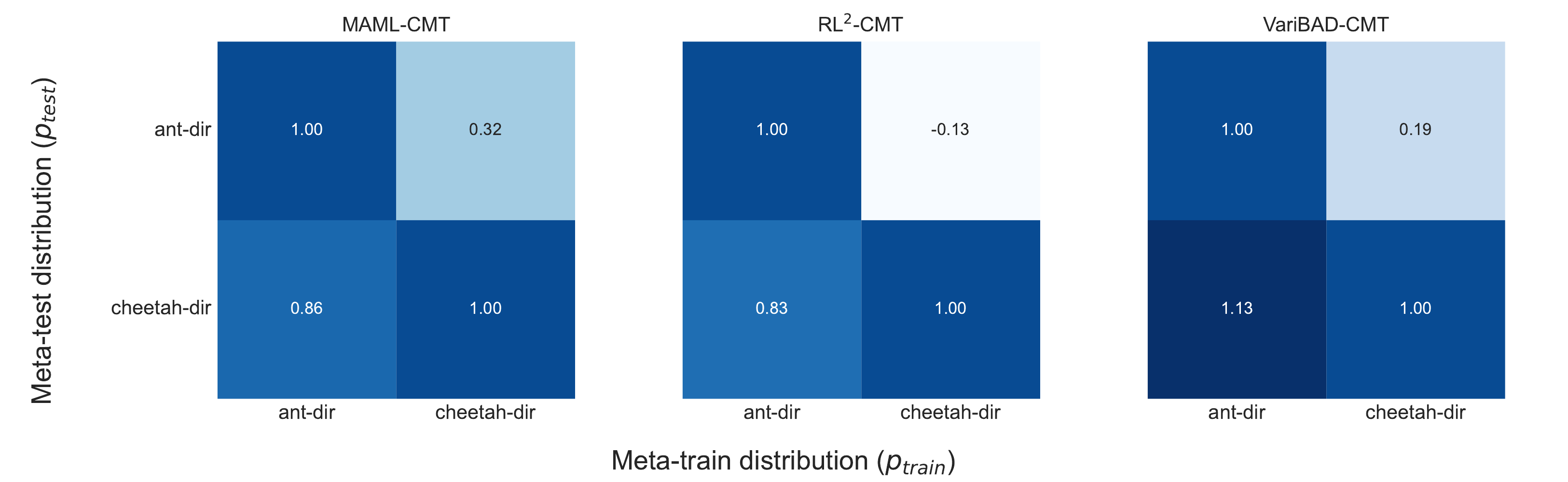}
    \caption{Consistency scores of different meta-RL algorithms on cheetah-direction $\leftrightarrow$ ant-direction; multi-task setting. 
    }
    \label{fig:ant-cheetah-dir-consistency-score}
\end{figure}

\begin{figure}[t]
    \centering
    \includegraphics[width=\textwidth]{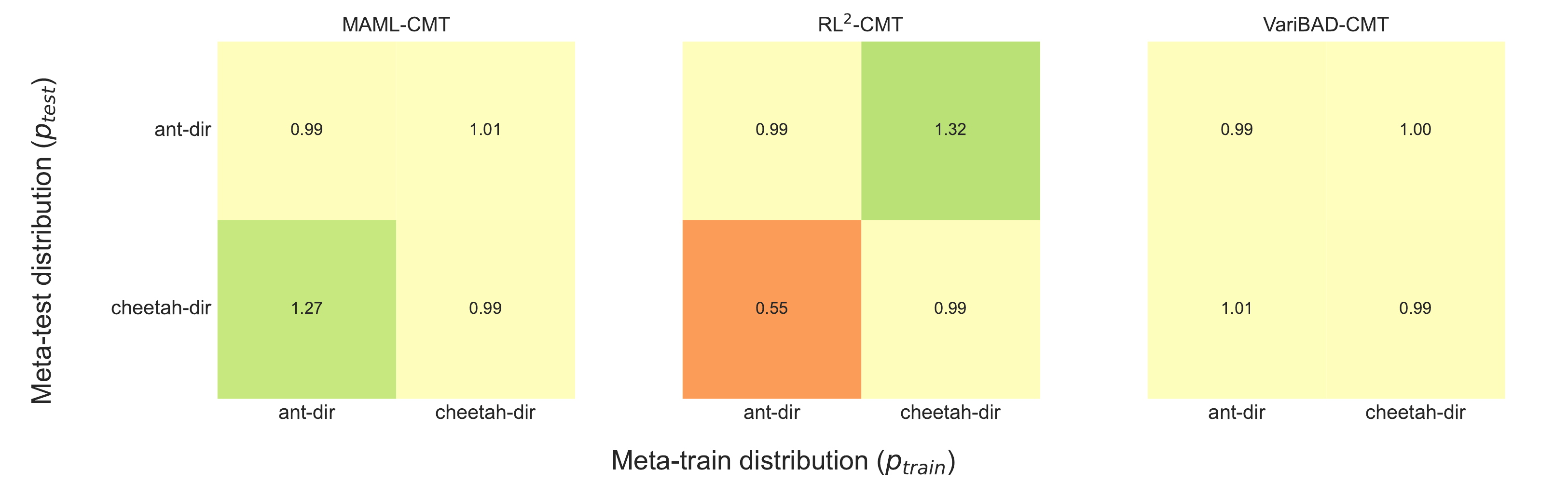}
    \caption{Consistency rates of different meta-RL algorithms on cheetah-direction $\leftrightarrow$ ant-direction; multi-task setting. 
    }
    \label{fig:ant-cheetah-dir-rate}
\end{figure}

\textbf{Cheetah direction $\leftrightarrow$ ant direction.} 
The consistency score and rate are shown in Figures \ref{fig:ant-cheetah-dir-consistency-score} and \ref{fig:ant-cheetah-dir-rate}. 
The transfer is asymmetric between the two tasks, similar to the trend under the single-task setting.  

\section{In-distribution Adaptation of Context-based Methods}
\label{appendix:adaptation-curve}

\begin{figure}[t]
    \centering
    \includegraphics[width=\textwidth]{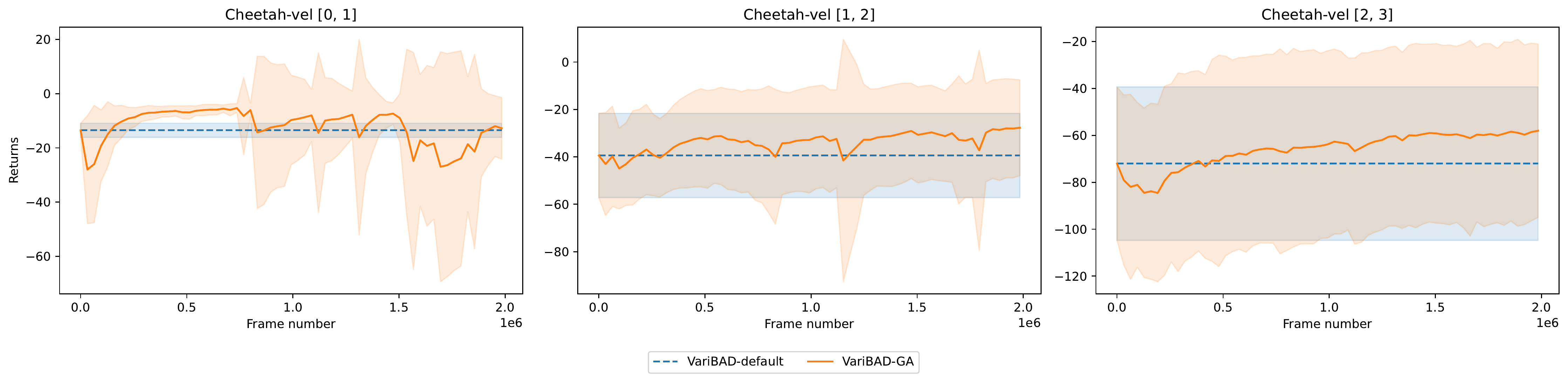}
    \caption{In-distribution adaptation curves of VariBAD on cheetah-velocity $[0,1]$, $[1,2]$, $[2,3]$. The {\color{blue}blue} dashed lines represent VariBAD's default adaptation performance. These lines are flat because context-based methods by default have no gradient update during adaptation. The {\color{orange}orange} lines represent how VariBAD's adaptation performance changes via gradient update. }
    \label{fig:varibad-cheetah-vel-in-dis-adaptation-curve}
\end{figure}

\begin{figure}[t]
    \centering
    \includegraphics[width=\textwidth]{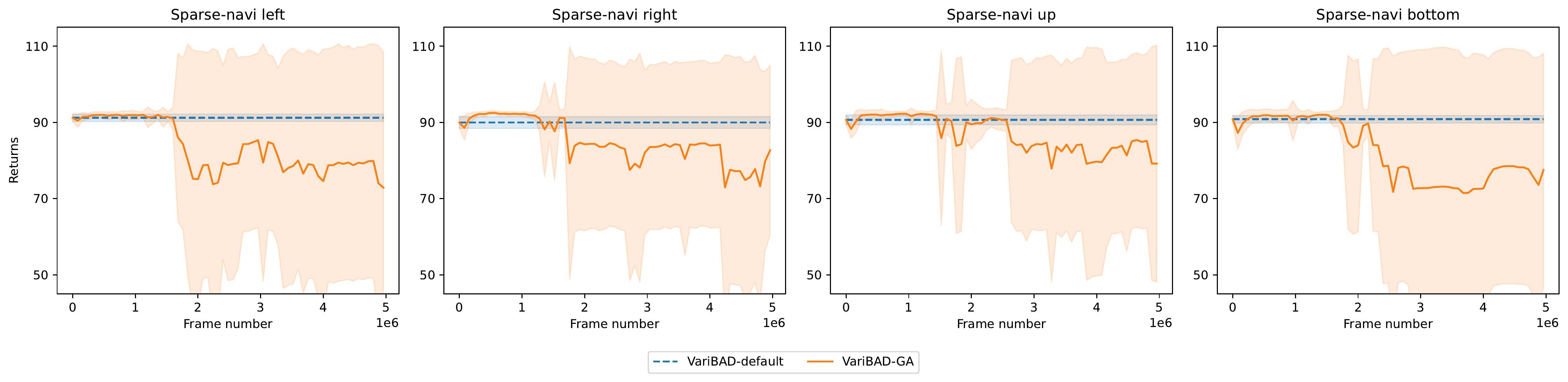}
    \caption{In-distribution adaptation curves of VariBAD on sparse-navigation to different directions.}
    \label{fig:varibad-sparse-navi-in-dis-adaptation-curve}
\end{figure}

We compare how context-based methods adapt to in-distribution test tasks via its default setting or gradient adaptation. 
For example, in-distribution adaptation curves of VariBAD on cheetah-velocity and sparse-navigation are shown in Figures \ref{fig:varibad-cheetah-vel-in-dis-adaptation-curve} and \ref{fig:varibad-sparse-navi-in-dis-adaptation-curve} respectively. 
On some tasks (especially sparse-navigation tasks), We can observe a dip in performance of VariBAD-GA, which indicates that it's less stable compared to the default adaptation setting without gradient update. 


\section{Behavior Analysis on Sparse Navigation}
\label{appendix:sparse-navigation-analysis}

As shown in Section \ref{subsec:result-single-task} and \ref{subsec:result-multi-task}, none of the meta-RL algorithms tested can consistently solve the navigation task with sparse reward under the OOD setting, but the context-based methods with gradient adaptation perform better than MAML. We thus look deeper into the behaviors of different methods during OOD adaptation to investigate what's the reason of this phenomenon. 

For MAML, we found that the gradients of all network parameters remain 0 during the whole adaptation process, thus it fails to learn anything during OOD adaptation. This is caused by MAML only exploring a very restricted area in the direction of the meta-training goal space, thus not collecting any trajectories with non-zero returns in the direction of the OOD goal space for policy update.

\begin{figure}[t]
    \centering
    \includegraphics[width=\textwidth]{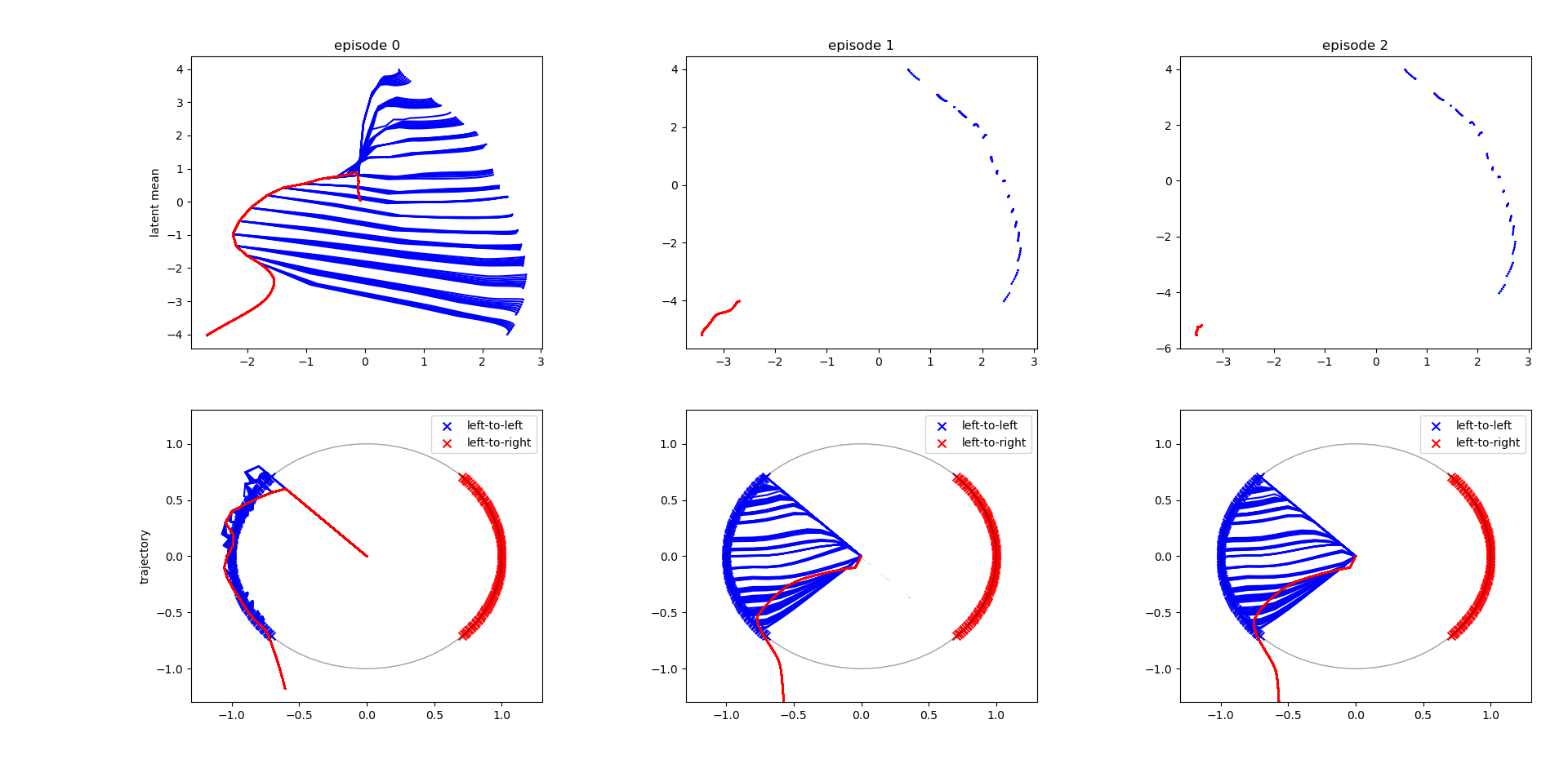}
    \caption{Visualization of VariBAD's behaviors on sparse navigation under its default adaptation setting. Each column shows the results of one episode within an RNN rollout. The first row shows how the inferred context with 2 dimensions evolves (starting from $[0, 0]$ in the first episode) within each episode. The second row shows the trajectory (starting from the origin) of the agent in each episode. The model is meta-trained on sparse navigation to left, and we uniformly sample 100 goals in both {\color{blue}left} and {\color{red}right} for test. 
    }
    \label{fig:varibad-sparse-navigation-default-adaptation}
\end{figure}

\begin{figure}[t]
    \centering
    \includegraphics[width=\textwidth]{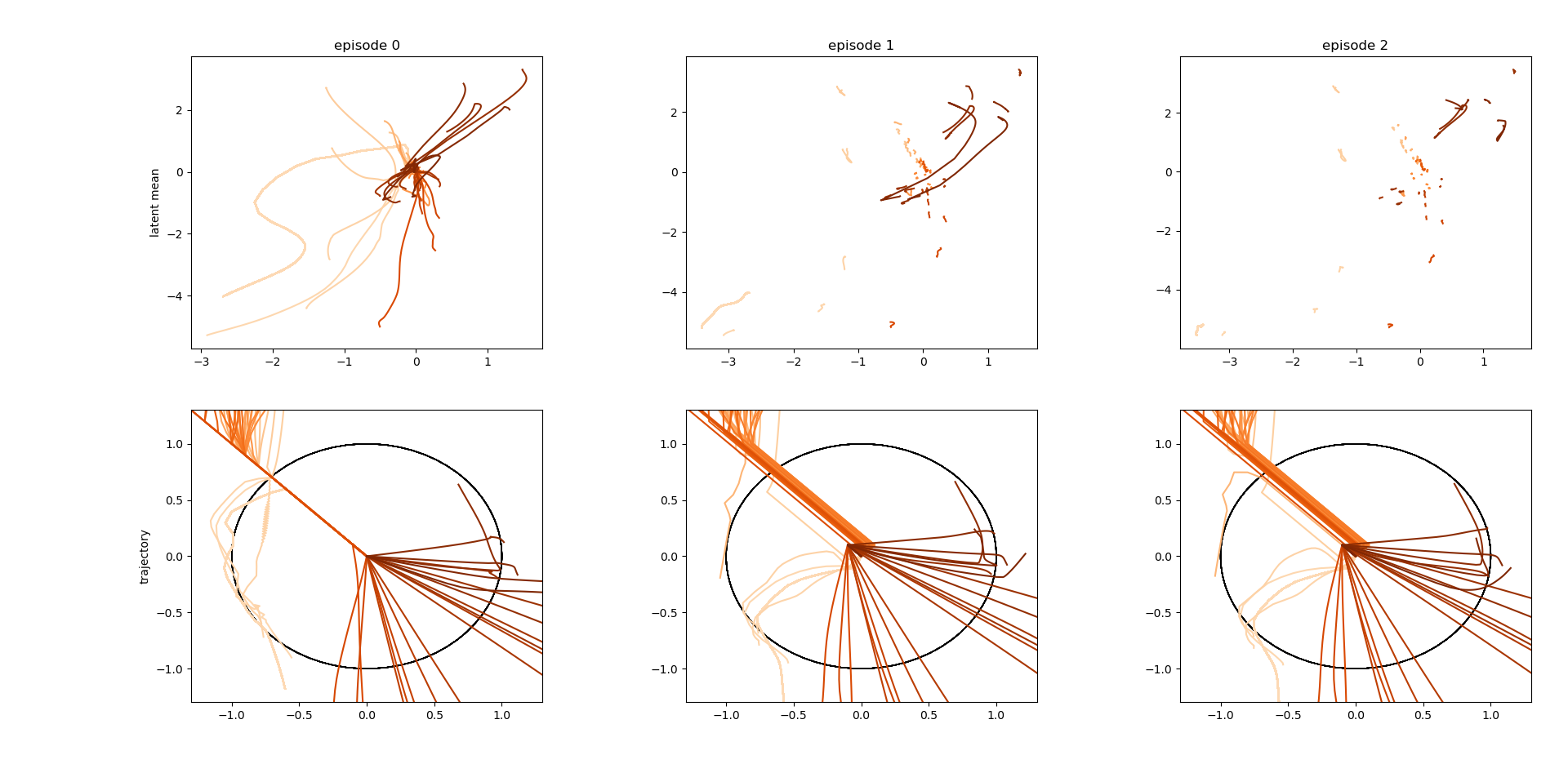}
    \caption{Visualization of VariBAD's behaviors on sparse navigation via gradient adaptation. Each column shows the results of one episode within an RNN rollout. The first row shows how the inferred context with 2 dimensions evolves within each episode. The second row shows the trajectory of the agent in each episode. The model is meta-trained on sparse navigation to left and adapted to right. The lighter color corresponds to the earlier stage during adaptation.
    }
    \label{fig:varibad-sparse-navigation-continued-training}
\end{figure}

On the contrary, we found that context-based methods explore better than MAML. Figure \ref{fig:varibad-sparse-navigation-default-adaptation} and \ref{fig:varibad-sparse-navigation-continued-training} show VariBAD's behaviors with its default adaptation setting and via gradient adaptation respectively, where a model is meta-trained to go left and adapted to go right. Under its default adaptation setting, VariBAD follows a fixed path to the left, and fails to explore other areas. However, as gradient update begins, VariBAD starts to randomly explore the space, and sometimes ends up in the right direction. 
Although VariBAD also just gets trajectories with zero returns before it finds the right direction during gradient adaptation, there are some components inside it that implicitly encourage exploration via parameter update: (1) the VAE is updated to fit to the new transition distribution, which can help the policy explore by providing new context input; (2) the critic module in the policy network is updated for better state value estimation, which encourages parameter update in the policy to try new behaviors. 
These components which implicitly encourages exploration are missing in MAML, which may explain why context-based methods can get much better performance than MAML on the sparse navigation task. 
However, as the effect of parameter update on the exploration behavior is quite random, we can not guarantee any convergence to the right goal direction, which may explain why they still can not stably match the final performance of in-distribution learning. 

\end{document}